\documentclass{article}

\usepackage{arxiv}
\usepackage{amssymb,amsmath,mathtools}
\usepackage{amsmath}
\usepackage{booktabs}
\usepackage{subcaption}
\usepackage{graphicx}
\usepackage{booktabs}
\usepackage{xfrac}
\usepackage{placeins}
\usepackage{array}
\usepackage{longtable,tabularx}
\usepackage{siunitx}
\usepackage{algorithm2e}
\usepackage{algpseudocodex}
\usepackage{hyperref}
\usepackage{natbib}
\usepackage[utf8]{inputenc}
\usepackage[T1]{fontenc}
\usepackage[export]{adjustbox}

\numberwithin{equation}{section}

\newcolumntype{P}[1]{>{\small}{p{#1}}}
\newcolumntype{L}[1]{>{\raggedright\let\newline\\\arraybackslash\hspace{0pt}}m{#1}}
\newcolumntype{C}[1]{>{\centering\let\newline\\\arraybackslash\hspace{0pt}}m{#1}}
\newcolumntype{R}[1]{>{\raggedleft\let\newline\\\arraybackslash\hspace{0pt}}m{#1}}

\newcommand{\mbf}[1]{\mathbf{#1}}
\newcommand{\mcf}[1]{\mathcal{#1}}

\newcommand{\x}[0]{\mathbf{x}}

\newcommand{\R}[0]{\mathbb{R}}
\newcommand{\N}[0]{\mathcal{N}}

\newcommand{\elbo}[0]{\mathcal{L_{\mathrm{ELBO}}}}
\newcommand{\gvn}[0]{\,|\,}
\newcommand{\E}[0]{\mathbb{E}}

\title{Gradient-enhanced deep Gaussian processes for
multifidelity modelling}

\author{
    \href{https://orcid.org/0000-0003-4012-4211}{\includegraphics[scale=0.06]{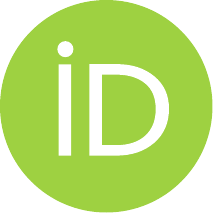}\hspace{1mm}Viv A.~Bone} \\
	Department of Electrical and Electronic Engineering\\
	The University of Melbourne\\
	Parkville, VIC, 3010 \\
	\texttt{viv.bone@unimelb.edu.au} \\
	\And
    Chris van der Heide \\
	Department of Electrical and Electronic Engineering\\
	The University of Melbourne\\
	Parkville, VIC, 3010 \\
	\texttt{chris.vanderheide@unimelb.edu.au} \\
	\And
    Kieran Mackle \\
	School of Mechanical and Mining Engineering\\
	The University of Queensland\\
	St Lucia, QLD, 4072 \\
	\texttt{m.kearney@uq.edu.au} \\
	\And
    \href{https://orcid.org/0000-0001-6962-8187}{\includegraphics[scale=0.06]{orcid.pdf}\hspace{1mm}Ingo H.\,J.~Jahn} \\
	School of Engineering\\
	The University of Southern Queensland\\
	Springfield, QLD, 4300 \\
	\texttt{ingo.jahn@unisq.edu.au} \\
    \And
    \href{https://orcid.org/0000-0002-9791-7776}{\includegraphics[scale=0.06]{orcid.pdf}\hspace{1mm}Peter M.~Dower} \\
	Department of Electrical and Electronic Engineering\\
	The University of Melbourne\\
	Parkville, VIC, 3010 \\
	\texttt{pdower@unimelb.edu.au} \\
    \And
    \href{https://orcid.org/0000-0002-9969-0982}{\includegraphics[scale=0.06]{orcid.pdf}\hspace{1mm}Chris Manzie} \\
	Department of Electrical and Electronic Engineering\\
	The University of Melbourne\\
	Parkville, VIC, 3010 \\
	\texttt{pdower@unimelb.edu.au} \\
}

\renewcommand{\shorttitle}{Gradient-enhanced deep Gaussian processes for
multifidelity modelling}

\hypersetup{
pdftitle={Gradient-enhanced deep Gaussian processes for multifidelity modelling},
pdfsubject={eess.SY},
pdfauthor={Viv A.~Bone, Chris van der Heide, Kieran Mackle, Ingo H.\,J.~Jahn, Peter M.~Dower, Chris Manzie},
pdfkeywords={gaussian process regression, deep gaussian process, gradient, computational fluid dynamics},
}

\begin{document}
\maketitle


\begin{abstract}
Multifidelity models integrate data from multiple sources to produce a single
approximator for the underlying process.
Dense low-fidelity samples are used to reduce interpolation error, while sparse
high-fidelity samples are used to compensate for bias or noise in the
low-fidelity samples.
Deep Gaussian processes (GPs) are attractive for multifidelity modelling as they
are non-parametric, robust to overfitting, perform well for small datasets, and,
critically, can capture nonlinear and input-dependent relationships between
data of different fidelities.
Many datasets naturally contain gradient data, especially when they are
generated by computational models that are compatible with automatic
differentiation or have adjoint solutions.
Principally, this work extends deep GPs to incorporate gradient data.
We demonstrate this method on an analytical test problem and a realistic partial
differential equation problem, where we predict the aerodynamic coefficients
of a hypersonic flight vehicle over a range of flight conditions and geometries.
In both examples, the gradient-enhanced deep GP outperforms a gradient-enhanced
linear GP model and their non-gradient-enhanced counterparts.
\end{abstract}


\section{Introduction}
\label{sec:introduction}

Across science and engineering, we often seek to model some underlying process
from sampled data.
This data is often available from several sources; accurate data is usually
expensive to obtain, while biased or noisy data is more readily available.
Multifidelity methods aim to fuse high- and low-fidelity data to produce a
single predictor
that is valid across all fidelities~\citep{peherstorfer2018survey}.
These methods use densely-sampled lower-fidelity data to reduce interpolation
error and sparsely-sampled higher-fidelity data to systematically compensate for
the bias and noise that corrupt the lower-fidelity data.
This approach can significantly increase the accuracy of the surrogate for a
given computational
budget~\citep{kennedy2000predicting,perdikaris2017nonlinear}.

Gaussian processes (GPs) are a family of stochastic processes that are
completely specified by their mean and covariance
functions~\cite{williams2006gaussian}.
Powerful and principled multifidelity
models~\citep{le2014recursive,rokita2018multifidelity,liu2018cope,forrester2007multi}
can be constructed by representing the latent function associated with each
fidelity level as a realization of GP.
GPs are attractive for regression and function approximation problems for
several reasons:
their probabilistic construction naturally yields uncertainty predictions;
they are non-parametric, so their complexity grows with the size of the dataset;
they are both highly flexible and robust to overfitting;
empirically, they perform well for small datasets; and
they can incorporate prior modelling assumptions and gradient
information~\citep{williams2006gaussian}.
These characteristics have led to increasing use of Gaussian process regression
(GPR) across many areas of science and
engineering~\citep{deisenroth2013gaussian,ray2019bayesian,deringer2021gaussian,gelfand2016spatial,wu2014gaussian,lukaczyk2015surrogate,brevault2020overview}.

Many existing GPR-based multifidelty
methods~\citep{helterbrand1994universal,goovaerts1998ordinary} are special cases
of the so-called linear model of coregionalization
(LMC)~\citep{bourgault1991multivariable,alvarez2012kernels},
which assumes that the output for each fidelity is represented by a linear
combination of some underlying latent functions.
The seminal `AR1' autoregressive model~\citep{kennedy2000predicting} follows a
similar approach,
but relates only successive fidelity levels directly using a bias correction
term.
For both these methods, the global linear relationship between fidelity levels
admits solutions governed by a single joint Gaussian process, facilitating
exact computation of the posterior predictions and the marginal
likelihood~\citep{alvarez2012kernels,kennedy2000predicting}.

However, the performance of classical multifidelity techniques degrades when
the relationship between fidelity levels is not linear or when the
correlation between fidelities varies across the input space.
The nonlinear autoregressive GP (NARGP)~\citep{perdikaris2017nonlinear} and
multifidelity deep GP~\citep{cutajar2019deep} were developed to address these
issues.
These methods generalize the AR1 model and assume that the outputs at fidelity
level $\ell$ are predicted by nonlinearly transforming the outputs from fidelity
level $\ell-1$ and adding a correction term that is a function of the original inputs.
NARGP further imposes that the training data are nested and that measurements
are noise-free, allowing the marginal likelihood to be explicitly
evaluated~\citep{perdikaris2017nonlinear}.
The multifidelity deep GP~\citep{cutajar2019deep} leverages recent advances in
sparse variational inference to relax these assumptions and construct an
approximate non-Gaussian posterior and marginal
likelihood~\citep{damianou2013deep,salimbeni2017doubly}.
This structure affords the multifidelity deep GP additional flexibility and
generality, but adds computational cost and cedes some theoretical properties of
standard GP models.

A further advantage of GPs is that they can naturally incorporate gradient
data, which leads to improved predictive accuracy and uncertainty
estimates~\citep{bouhlel2019gradient,han2013improving,lukaczyk2015surrogate}.
Gradient data is particularly common when approximating the output of an
expensive high-fidelity computational model from a small set of data.
In this setting, gradient data can often be cheaply and conveniently obtained
via automatic differentiation tools~\citep{baydin2018automatic}.
Moreover, when considering solutions to discretized systems of differential
equations, cheap approximate gradients can instead be obtained with lower memory
consumption via adjoint
methods~\citep{jameson1988aerodynamic,nadarajah2000comparison}.
Multifidelity data also naturally arises in this context, often corresponding to
increasing levels of simulation accuracy~\citep{peherstorfer2018survey}.
When simulating physical systems, data of different fidelities can be generated
by employing models with different levels of simplifying assumptions.
For example, in computational fluid dynamics, these fidelity levels might
correspond to inviscid simulations, Reynolds averaged Navier Stokes simulations,
and large eddy simulations~\citep{versteeg2007introduction} (in order of
increasing accuracy).
Moreover, when simulating any system of discretized partial differential
equations (PDEs), multifidelity data may be generated by utilizing different
mesh discretizations~\citep{peherstorfer2018survey}.

While classical multifidelity GPR techniques (i.e.,\, those that assume linear
relationships between fidelity levels) can be routinely extended to incorporate
gradient data, this extension is less clear for the modern multifidelity
approaches.
Gradient data has been incorporated into a nonlinear multifidelity kriging method,
but this method assumes a specific form for the bridging function that relates
fidelity levels~\citep{han2013improving}.
This work focuses on extending the multifidelity deep GP, which is a more
general method, to incorporate gradient information.
We first illustrate this technique on the multifidelity branin function,
which is a common analytic test problem~\citep{perdikaris2017nonlinear},
then we consider a challenging aerospace application, where we predict the
aerodynamic coefficients of hypersonic flight vehicle over a range of flight
conditions and vehicle geometries.
The techniques presented herein also apply to deep GPs trained on only
single-fidelity data.

This paper is structured as follows:
Sec.\;\ref{sec:gaussian_process_regression} provides relevant background on GPR,
Sec.\;\ref{sec:gradient_enhanced_deep_gp} presents the gradient-enhanced deep GP,
Sec.\;\ref{sec:implementation} summarizes the implementation of the GPR models,
Sec.\;\ref{sec:results_test_problem} presents numerical results on an analytical
test problem,
Sec.\;\ref{sec:results_pde_problem} treats the representative aerospace test problem,
and
Sec.\;\ref{sec:conclusion} concludes the paper.

\section{Review of Gaussian process regression}
\label{sec:gaussian_process_regression}

This section reviews the elements of GPR theory that are relevant to the
construction of our method, including background to GPs, gradient-enhancement,
sparse variational inference, and deep GPs.
A GP is a stochastic process $f$ on $\R^d$ characterized by the property that
given any finite collection of points $x_1,\dots,x_n\in\R^d$, the vector
$[f(x_1),\dots,f(x_n)]^\intercal$ has a multivariate Gaussian distribution.
The mean and covariance functions $m:\R^d\to\R$  and $k:\R^d\times\R^d\to\R$
completely specify the GP: if, given $x,x^\prime\in\R^d$, we have $\mathbb{E}
f(x) = m(x)$ and $\mathbb{E}(f(x) - m(x))(f(x^\prime) - m(x^\prime)) =
k(x,x^\prime)$, that is $(f(x_1),\dots,f(x_n))\sim N(m_{X},K)$ where $(m_X)_i =
m(x_i)$ and $K_{ij} = k(x_i,x_j)$ for $i,j=1,\dots,n$, then we say that
$f\sim\mathcal{GP}(m,k)$.
We assume that the kernel is Gaussian with variance and lengthscale parameters
$\sigma^2$ and $l_i$ for $i=1,...,d$, i.e.  $k(x,x^\prime)\doteq
\sigma^2\exp\left(\frac{1}{2}\sum_{i=1}^d|x_i - x_i^\prime|^2/l_i\right)$.
Other choices of kernel can be used in our construction provided they are $C^2$.

When using GPs for regression tasks, inference is performed using Bayes' rule.
Given a set $(X_i,Y_i)_{i=1}^{n}$ of independent and identically distributed
input-output data pairs, the assumption that each $Y_i = \tilde{f}(X_i) +
\varepsilon_i$ for a latent function $\tilde{f}$ with $\varepsilon_i\sim
N(0,\sigma_n)$ induces a Gaussian likelihood.
That is, $p(Y \gvn f,X) =
(2\pi\sigma^2_n)^{-n/2}\exp(-\frac{1}{2\sigma^2}\|Y-\tilde{f}(X)\|^2)$, with
$\sigma^2_n$ capturing model and output noise.
While the case where $m\equiv0$ is most commonly considered, the hierarchical
structure of both multifidelity models and deep GPs admit natural prior means.
The covariance structure is encoded in the gram matrix $K$, which has elements
$K_{ij} = k(X_i,X_j)$.
When considering a collection of $N$ new input points $X^\star =
\{x_1^\star,\dots,x_N^\star\}$, we write $f(X^\star) =
(f(x_1^\star),\dots,f(x_N^\star))$ and denote by $K^{\star\star}\in\R^{N\times
N}$ and $k^{\star}\in \R^{n\times N}$ the matrices with entries
$K^{\star\star}_{ij} = k(x_i^\star,x_j^\star)$ and $K^{\star}_{ij} =
k(X_i,x_j^\star)$.
In this construction, our choice of GP prior $f\sim \mathcal{GP}(m,k)$
conditioned on the data specifies a posterior distribution $p(f\,|\, X,Y)$ that
satisfies~\cite{williams2006gaussian}
$f(X^\star)\,|\, X,Y\sim \N(f^\star(X^\star),\Sigma(X^\star))$, where
\begin{align}
\label{eq:posterior_mk}
f^\star(X^\star) &= m_{X^\star}
- K^{\star\top}(K + \sigma^2_n I)^{-1}(Y - m_X), \quad \text{for} \quad (m_{X^\star})_i\doteq
m(x^\star_i)
\\ \nonumber
\Sigma(X^\star) &= K^{\star\star} - K^{\star\top}(K + \sigma^2_n I)^{-1}K^\star
\end{align}
and $Y$ denotes the vector with $i$-th component $Y_i$.
The model hyperparameters $\theta$, which include kernel parameters
$l_1,\, ...,\, l_d$ and noise variance $\sigma_n^2$, are then typically
optimized in an empirical Bayes procedure to (locally) maximize the log-marginal
likelihood
\begin{align}\label{eq:log_ml}
    L(Y\,|\, X,\theta) = -Y^\top(K + \sigma^2_n I)^{-1}Y - \frac{1}{2}\log|K + \sigma^2_n I| - \frac{n}{2}\log(2\pi),
\end{align}
which is equivalent to the free energy principle. This Bayesian approach inherently balances model complexity and prediction
accuracy; the first term in \eqref{eq:log_ml} penalizes prediction error while the second penalizes model complexity.

\subsection{Gradient enhancement}
\label{sec:gradient_enhancement}

Many datasets in the physical sciences and engineering, especially when generated with computational models, contain gradient information in addition to output values.
This information can be incorporated into GP models to improve their
prediction accuracy and uncertainty estimates~\citep{williams2006gaussian}.
We consider datasets where the output samples are augmented with gradients ---
i.e.,\, $(Y_\nabla)_i = [Y_i,\nabla^\intercal Y_i]^\intercal$ --- where $\nabla Y$ denotes the $d$-dimensional sample of gradient information.
We assume independent noise on each derivative measurement~\citep{lukaczyk2015surrogate}, so each $(Y_{\nabla})_{i} \in \R^{1 + d}$ is a sample from
\begin{equation}
\label{eq:grad_outputs}
\mcf{N}\left(
    \begin{bmatrix} \tilde{f}(X_i) \\  \nabla \tilde{f}(X_i)
    \end{bmatrix},\,
    \text{diag}\left(\sigma_{1}^2,\, ...,\, \sigma_{d+1}^2\right)
\right).
\end{equation}
Since the kernel is assumed to be smooth,
joint predictions for $f$ and $\nabla f$ can be generated using
using the gradient kernel
$k_{\nabla} : \R^{d} \times \R^{d} \to \R^{(d+1)\times(d+1)}$
\begin{equation}
\label{eq:grad_k}
k_\nabla(x_p,\, x_q) =
    \begin{bmatrix}
    k(x_p,\, x_q) & \nabla_{q}^\intercal k(x_p,\, x_q)
    \\
    \nabla_{p} k(x_p,\, x_q) &
        \nabla_{p} \nabla_{q}^\intercal k(x_p,\, x_q)
    \end{bmatrix},
\end{equation}
for all $x_p,\, x_q \in \R^d$,
where $\nabla_p$ ($\nabla_q$) denotes the derivative taken with respect to
$x_p$ ($x_q$), and $k_\nabla(x_p,x_q)$ is the covariance of $f_\nabla(x_p)$ and
$f_\nabla(x_q)$ for $f_\nabla \doteq [f,\nabla^\top f]^\top$. The mean vector of
$f_\nabla$ is denoted $m_\nabla \doteq [m,\nabla^\top m]^\top$. The form of
\eqref{eq:grad_k} follows from linearity of differentiation and is positive
definite for stationary kernels -- for details see~\citep{williams2006gaussian}.
The gradient kernel can be used to generate predictions in exactly the same way as the gradient-free case, with each entry of $K$, $K^\star$ and $K^{\star\star}$ being replaced by blocks of the form given in \eqref{eq:grad_k}.
Samples from $f_\nabla\,|\,Y_\nabla,X$
now correspond to selecting
only functions that (approximately) pass through the data at the observed gradients.
This process improves the predictions generated by the GP, particularly for sparse training data.

\subsection{The linear model of coregionalization}
\label{sec:multifidelity_methods_lmc}

Multifidelity GPR methods are data fusion techniques that approximate a family of related latent functions
$\tilde{f}^1$, ..., $\tilde{f}^L$ from iid samples that contain information from
each of the $L$ fidelities,
$(X^\ell_i,Y^\ell_i)_{\ell,i=1}^{L,n^\ell}$,
where $n^\ell$ is the number of datapoints for fidelity $\ell$.
Many classical multifidelity techniques are special cases of the LMC scheme,
which models each output as a linear combination of some underlying latent functions~\citep{bourgault1991multivariable}.
This assumption induces a product kernel which is separable in its vector-valued point-wise arguments
$x_p$, $x_q$ and integer-valued layer-wise indices $i$, $j$,
which represent model fidelities:
\begin{equation}
\text{cov}(f^i(x_p),\, f^j(x_p)) = k(x_p,\, x_q) \cdot k_\mathrm{I}(i,\, j)
\end{equation}
where $k : \R^d \times \R^d \to \R$ is a standard (or gradient-enhanced) GP kernel and
$k_\mathrm{I} : \mathbb{N} \times \mathbb{N} \to \R$ is an index kernel.
The kernel matrix $K_I$ corresponding to the kernel $k_\mathrm{I}$ is positive
semidefinite, typically parameterized as a Cholesky decomposition, i.e., $k_\mathrm{I}(i,\, j) = \left[BB^{\intercal} \right]_{ij}$, where $B \in \R^{L\times L}$ is an upper triangular matrix, and the $i,j$-th entry gives the scaling of the shared kernel between model fidelities $i$ and $j$.
LMC can be extended to consider a sum of $T$ separable kernels, each with
different hyperparameters:
\begin{equation}
\label{eq:lmc_cov}
\text{cov}(f^i(x_p),\, f^j(x_p))
= \sum_{t=1}^T
    \left( k_t(x_p,\, x_q) \cdot {k_\mathrm{I}}_t(i,\, j) \right).
\end{equation}
The common `AR1' approach~\citep{kennedy2000predicting} can be shown to be a
special case of LMC with number of separable kernels set to the number of
fidelities (i.e., $T=L$).

LMC models all the latent functions, each of which corresponds to a different fidelity level, as a single multi-output GP
$f_{MF} = [f^1,\, ..., f^L]^\top$. 
For a finite collections of points, the gram matrix is
\begin{align}
\label{eq:k_lmc}
K_{LMC} =
\begin{bmatrix}
K_{11} & ... & K_{1L}
\\
\vdots & \ddots & \vdots
\\
K_{L1} & ... & K_{LL}
\end{bmatrix},
\end{align}
where each submatrix $K_{ij}$ is the kernel matrix evaluated with 
the covariance function in \eqref{eq:lmc_cov} for input sets $X^i$ and $X^j$ and
indices $i$ and $j$ for $i,j \in 1,...,L$.
We remark that in the case where training data for all fidelity levels contain every input point, this can be computed as the Kronecker product $K_I\otimes K$, where $(K_I)_{i,j} = k_I(i,j)$ and $K$ is the gram matrix with entries $K_{p,q} = k(X_p,X_q)$.
LMC can naturally be extended to consider gradient information by using gradient kernels  from \eqref{eq:grad_k} for each $k_t$.
Assuming equal measurement noise $\sigma_n^2$ for each fidelity level, the
log-marginal likelihood~\eqref{eq:log_ml} and posterior predictive mean and
covariance terms~\eqref{eq:posterior_mk} can be evaluated in the same way as the single-fidelity case.

\subsection{Multifidelity deep Gaussian processes}
\label{sec:multifidelity_deep_gaussian_processes}

An emerging technique that can capture nonlinear relationships between different model fidelities is the multifidelity deep GP~\citep{cutajar2019deep}.
The deep GP defines the prior recursively, with the inputs to layer
$\ell$ being the outputs from layer $\ell-1$, augmented with the base-layer 
input~\citep{damianou2013deep}.
Under this construction, layer one is a standard GP with deterministic inputs, then from the second layer
onwards, we assume the model has the following autoregressive form \citep{cutajar2019deep}:
\begin{equation}
f^\ell(x) = g^\ell(f^{\ell-1}(x),\, x) + \gamma^\ell(x),
\end{equation}
where $g^\ell : \R \times \R^d \to \R$ and $\gamma^\ell : \R^d \to \R$.
To represent this model as a GP, we assume that the mapping
$g^\ell$ decomposes as \citep{cutajar2019deep}
\begin{equation}
\label{eq:dgp_regressive_form}
f^\ell(x) = g_f^\ell(f^{\ell-1}(x)) \cdot g_x^\ell(x) + \gamma^\ell(x).
\end{equation}
Note that \eqref{eq:dgp_regressive_form} can be represented by combining 
three GPs with mean functions
$m^\ell_{gx} : \R^d \to \R$,
$m^\ell_{g f} : \R \to \R$,
$m^\ell_{\gamma x} : \R^d \to \R$
and covariance functions
$k^\ell_{gx} : \R^d \times \R^d \to \R$,
$k^\ell_{g f} : \R \times \R \to \R$,
$k^\ell_{\gamma x} : \R^d \times \R^d \to \R$. 
The compositional structure of the deep GP results in a model that is not a GP
itself, since its values evaluated at finitely many points cannot be described
by a multivariate normal distribution~\citep{damianou2013deep}.

We set $g^\ell_{x}$ to be everywhere one and assume a zero mean prior for $m^\ell_{\gamma x}$,
reducing the mean function $m^\ell : \R \times \R^d \to \R$ to
$m^\ell(f^{l-1}(x),\, x) = m^\ell_{g f}(f^{l-1}(x)) = \mathbb{E}f^{\ell}(x)$.
We then choose an affine mean for $m^\ell_{g f}$, i.e.,
\begin{align}
\label{eq:affine_mean}
m^\ell(f^{l-1}(x),\, x) = m^\ell_{g f}(f^{\ell-1}(x)) &\doteq \kappa f^{\ell-1}(x) + c,
\end{align}
with the learned parameters $\kappa$ and $c$ initialzed to unity and zero respectively. 
That is, as a prior, we assume the outputs for fidelity level $\ell$ are equal to
those for fidelity level $\ell-1$. 
The layer-wise covariance functions are of the form
\begin{align}
\label{eq:dgp_covar}
\text{cov}(f^\ell(x_p),\, f^\ell(x_q))
    &= \mathbb{E}[
    (f^\ell(x_p) - m^\ell(f^{\ell-1}(x_p),\, x_p))(f^\ell(x_q) - m^\ell(f^{\ell-1}(x_q),\, x_q))]
\\ \nonumber
    &\doteq k^\ell((f^{\ell-1}_p,\, x_p),\, (f^{\ell-1}_q,\, x_q)),
\end{align}
where we write $f^{\ell-1}_p = f^{\ell-1}(x_p)$, $f^{\ell-1}_q = f^{\ell-1}(x_q)$,
and $k^\ell : \R \times \R^d \times \R \times \R^d \to \R$.
Following \citep{salimbeni2017doubly},
in each layer's covariance function, we include internal noise (which propagates through all subsequent layers), giving
\begin{equation}
\label{eq:dgp_kernel_form}
k^\ell((f^{\ell-1}_p,\, x_p),\, (f^{\ell-1}_q,\, x_q)) = 
    k^\ell_{gx}(x_p,\, x_q) \cdot k^\ell_{g f}(f^{\ell-1}_p,\, f^{\ell-1}_q)
    + k^\ell_{\gamma x}(x_p,\, x_q) + \sigma_k^2\, \delta_{p,q}
\end{equation}
where $\sigma_k^2$ is the variance of the kernel noise, and $\delta_{p,q}$ is the Kronecker delta.
We  write $X^\ell = (X^\ell_i)_{i=1}^n$ and fix the first-layer inputs as the union of all input data $X^0 = \bigcup_{\ell=1}^L X^\ell$.
In contrast to the NARGP \citep{perdikaris2017nonlinear}, this approach does not require training data to be nested across fidelity levels ($X^1 \supseteq X^2 \supseteq ... \supseteq X^L)$.

We have seen that as observed in \cite{damianou2013deep}, the compositional
nature of deep GPs results in a stochastic process that is not itself a GP.
Since the prior is no longer Gaussian, the posterior distribution and marginal
likelihood are no longer analytically or computationally tractable.
However, in order to perform inference using these models, variational
approximations can be used.

\subsection{Variational inference}
\label{eq:variational_inference}

The fundamental challenge when fitting deep GPs is the evaluation of the marginal
likelihood and posterior distribution.
Direct numerical approximation of the marginal likelihood via Monte
Carlo sampling is only tractable for extremely small problems.
However, deep GPs can be made computationally tractable using variational inference (VI) techniques~\citep{damianou2013deep, salimbeni2017doubly}, which were 
previously used to scale GP models to large datasets~\citep{titsias2010bayesian}. 
These so-called sparse VI methods mitigate the $\mathcal{O}({n^3})$ computational bottleneck of inference by circumventing the need to invert the full gram matrix and by enabling subsampling. 

Sparse VI techniques introduce a family of \emph{inducing inputs} $Z = \{z_i\}_{i=1}^{m}$ with corresponding inducing points $U = \{u\}_{i=1}^m$ with each $z_i\in\R^d$ and $u_i\in\R$ assumed to satisfy $u_i = f(z_i)$. The joint density $p(f,U)$ is then a Gaussian with a common prior mean and covariance function. Thus, the joint distribution of $Y,f$ and $U$ decompose into the prior and likelihood, giving
\begin{align}\label{eqn:gppost}
    p(Y,f,U) = p(f\,|\, U;X,Z)p(U;Z)\prod_{i=1}^np(Y_i\,|\, f_i),
\end{align}
where we have factored the joint prior $p(Y,f,U\,|\,X,Z)$ as $U\sim\mathcal{N}(m_Z,K_{ZZ})$ and conditional likelihood $f\,|\, U \sim \mathcal{N}(\tilde\mu,\tilde\Sigma)$. Here, the mean and covariance of $U$ satisfy $(m_Z)_i\doteq m(z_i)$ and $(K_{ZZ})_{ij}\doteq k(z_i,z_j)$.
The conditional mean and covariance $\tilde\mu$ and $\tilde\Sigma$ are given by
\begin{align*}
    \tilde\mu_i \doteq m(X_i) + \alpha(X_i)^\top(U - m_Z)
    \qquad\text{and}\qquad
    \tilde\Sigma_{ij} \doteq  k(X_i,X_j) - \alpha(X_i)^\top K_{ZZ}\alpha(X_j),
\end{align*}
for $i,j=1,\dots,n$ and $\alpha(X_i)\doteq K_{ZZ}^{-1}K_{ZX_i}$,
where $(K_{ZX_i})_j = k(z_j, X_i)$.

For training, a lower bound on the marginal likelihood (the evidence lower bound, or `ELBO') is introduced as a surrogate cost in lieu of computing the log-marginal likelihood.
This cost is derived by introducing a free-form variational posterior
\begin{equation}
    q(f,U) \approx p(f,U\gvn Y).
\end{equation}
Writing the marginal likelihood as $p(Y) = \frac{p(f,U,Y)}{p(f,U\,|\, Y)}$ (where dependence on $X$ and $U$ is suppressed), then taking the log and expectation over $q(f,U)$, gives
\begin{align*}
    \log p(Y) &= \E_{q(f,U)}\log\left(\frac{p(f,U,Y)}{q(f,U)}\right) + \E_{q(f,U)}\log\left(\frac{q(f,U)}{p(f,U\gvn Y)}\right) \\
    &\doteq \elbo\left(q(f,U)\right) + \mathcal{D_{\mathrm{KL}}}\left(q(f,U)\,\|\,p(f,U\,|\, Y)\right)\geq \elbo\left(q(f,U)\right)
\end{align*}
where $\mathcal{D_{\mathrm{KL}}}$ denotes the (non-negative) Kullback-Leibler divergence (KL) from $q(f,U)$ to $p(f,U\gvn Y)$.
Exploiting \eqref{eqn:gppost}, the ELBO can be decomposed and marginalized to obtain
\begin{align}\label{eqn:elbo}
    \elbo\left(q(f,U)\right) &= \E_{q(f,U)}\log\left(\frac{p(f,U,Y)}{q(f,U)}\right)
     = \E_{q(f,U)}\log\left(\frac{p(Y\gvn f,U)p(f\gvn U)p(U)}{p(f\gvn U)q(U)}\right)\nonumber\\
     &= \E_{q(f)}\log p(Y\gvn f) - \mathcal{D_{\mathrm{KL}}}\left(q(U)\,\|\,p(U)\right).
\end{align}
For analytic and computational tractability, we set the form of the variational posterior to be
$q(f,U) = p(f\gvn U )q(U)$, where $q(U) = \mathcal{N}(m_q,S_q)$ is a Gaussian with variational parameters $m_q\in\R^m$ and $S_q\in\R^{m\times m}$~\citep{salimbeni2017doubly}.
Maximization of the ELBO can be viewed as maximization of the expected marginal likelihood under the choice of variational posterior, penalized by the KL divergence $q(U)$ to the true prior mean.
As the variational posterior is a product of two Gaussians, $U$ can be marginalized out analytically to obtain the Gaussian predictive distribution $q(f) = \mathcal{N}(\mu,\Sigma)$ with mean and covariance given by
\begin{align*}
    \mu_i \doteq m(X_i) + \alpha(X_i)^\top(m_q - m_Z)
    \qquad\text{and}\qquad
    \Sigma_{ij} \doteq  k(X_i,X_j) - \alpha(X_i)^\top \left(K_{ZZ} - S_\phi\right)\alpha(X_j).
\end{align*}

Inducing point methods can also be used to define a variational posterior for
deep GPs by introducing inducing inputs
$Z^\ell = \left\{z^\ell_i\right\}_{i=1}^{m_\ell}$ and inducing variables
$\{u^\ell\}_{i=1}^{m_\ell}$ for
each layer.
The inducing inputs correspond to each layer's inputs, so for the first layer
each $z_i^1 \in \R^d$, and for subsequent layers, the inducing inputs $z_i^\ell$
match the previous layer's output dimension. 
In the multifidelity setting, we consider single-width hidden layers, so each $z^\ell_i \in \R$ for $l\geq 2$ has a corresponding base-layer inducing input $z^1_i$.
For moderately sized datasets, we choose a full-rank variational posterior where $m_\ell = n_\ell$.
In this case, we fix the base-layer inducing inputs at the input data $Z^1 = X^0$
and set the remaining inducing inputs as the output data for the previous
layer, so
$z^\ell_i = \tilde{y}^{\ell-1}_i \approx \tilde{f}^{\ell-1}(x_i)$
for $l = 2,\, ...,\, L$ \citep{cutajar2019deep}.
For larger datasets we set
$m_\ell \leq n_\ell$ and including the inducing inputs as free variational parameters.
To construct an analogous  ELBO for a deep GP with $L$ layers, we suppress dependence on data and intermediate layers' internal dimensions to decompose the joint density as the likelihood and prior
\begin{align*}
    p(Y,\{f^\ell,U^\ell\}_{\ell=1}^L) \doteq \prod_{i=1}^n p(Y_i\gvn f^L_i)
    \prod_{\ell=1}^L p(f^\ell\gvn U^\ell;f^{\ell-1},Z^{\ell})p(U^\ell;Z^{\ell}),
\end{align*}
which can be compared with \eqref{eqn:gppost}. 

Doubly stochastic variational inference can be used to compute approximate deep Gaussian process posteriors that maintain correlations both within and between layers \citep{salimbeni2017doubly}.
Under this setup, the variational distribution of $U^\ell$
factorizes between layers as a Gaussian with mean $m_\phi^\ell$ and covariance
$S_{\phi}^\ell$, admitting a variational posterior of the form
\begin{align}\label{eq:variational_posterior}
q\Big(\big\{f^\ell,U^\ell\big\}_{\ell=1}^L\Big)\doteq \prod_{\ell=1}^L p\big(f^\ell\gvn U^\ell;f^{\ell-1},Z^{\ell}\big)q\big(U^\ell\big).
\end{align}
Since the distributions are Gaussian, the inducing variables can be marginalized out to obtain a Gaussian predictive distribution
\begin{align}
\label{eq:dgp_fl}
    q(\{f^\ell\}_{\ell=1}^L)\doteq \prod_{\ell=1}^L q(f^\ell\gvn m_\phi^\ell, S_\phi^\ell;f^{\ell-1},Z^{\ell}) = \prod_{\ell=1}^L\N(\mu^{\ell},\Sigma^{\ell}),
\end{align}
whose mean vectors and covariance matrices have entries
\begin{align}
\label{eq:dgp_mu_k}
    \mu_i^{\ell} &\doteq m^\ell(f^{l-1}_i, X_i) + \alpha^\ell(f^{\ell-1}_i, X_i)^\top(m^{\ell}_q - m^{\ell}_{Z})
\\ \nonumber
    \Sigma_{ij}^{\ell} &\doteq  k^\ell(f^{\ell-1}_i,f^{\ell-1}_j) -
\alpha^\ell(f^{\ell-1}_i, X_i)^\top \big(K_{ZZ}^{\ell} - S_q\big)\alpha^\ell(f^{\ell-1}_j, X_j),
\end{align}
where
$(m_Z^{\ell})_i = m^\ell(Z^\ell_i, Z^1_i)$ for the mean function \eqref{eq:affine_mean},
$(K_{ZZ}^\ell)_{ij} = k^\ell((Z^\ell_i, Z^1_i), (Z^\ell_j, Z^1_j))$ for the
covariance function \eqref{eq:dgp_covar},
and
$\alpha^\ell(f^{\ell-1}_i, x_i)\doteq (K^\ell_{ZZ})^{-1}K^\ell_{ZX_i}$
with 
$(K_{ZX_i}^\ell)_{j} = k^\ell((Z^\ell_j, Z^1_j), (f^{\ell-1}_i, x_i))$.
The $i$-th marginal of any layer $M$ of the posterior depends only on the $i$-th marginal of the previous layers, that is,
\begin{align}
\label{eq:qfMi}
    q(f^M_i) =  \int\prod_{\ell=1}^{M-1} q(f^\ell_i\gvn m_\phi^\ell, S_\phi^\ell;f^{\ell-1}_i,Z^{\ell})\,df^\ell_i,
\end{align}
which enables tractable computation of the posterior.
The integral appearing in \eqref{eq:qfMi} can be approximated by first sampling
$\varepsilon^\ell_i\sim\N(0,I_{d^\ell})$, then recursively drawing
\begin{equation}
\label{eq:q_sampling}
\hat{f}_{i}^\ell\sim q(f_{i}^\ell\gvn m^\ell, S^\ell; f^{\ell-1}_i,Z^{\ell})
\end{equation}
for $\ell = 1,\dots,L-1$ via $\hat{f}^\ell_i = \mu^{\ell}_i +
\varepsilon^\ell_{i}\odot\sqrt{\Sigma^{\ell}_{ii}}$, where $d^\ell$ is the
output dimension of the $\ell$-th layer, and $\hat{f}^0_{i} = X_i$.
This process is equivalent to representing each layer's posterior as a mixture of 
$N_s$ Gaussian distributions $\{\hat{q}^j(f^\ell)\}_{j=1}^{N_s}$,
where each distribution corresponds to a sample
$(\hat{f}^{\ell-1})^j = \{(\hat{f}^{\ell-1}_i)^j\}_{i=1}^{n^l}$ drawn
from the previous layer's corresponding output distribution, i.e.,
\begin{equation}
\hat{q}^j(f^{\ell}) = q(f^\ell_i\gvn m_\phi^\ell, S_\phi^\ell;\hat{f}^{\ell-1}_i,Z^{\ell}).
\end{equation}
Thus, the posterior at layer $\ell$ is approximated by the Gaussian mixture
\begin{equation}
\label{eq:approx_q}
q(f^\ell) \approx \hat{q}(f^\ell) \doteq \frac{1}{N_s} \sum_{j=1}^{N_s} \hat{q}^j(f^{\ell}),
\end{equation}
where samples can be drawn from $\hat{q}^j(f_i^{\ell})$
for $i = 1,...,n^\ell$ independently.
We denote the full set of samples over all layers as
$\hat{f} \doteq \{(\hat{f}^\ell)^j\}_{\ell,j=1}^{L,N_s}$
and the corresponding approximate posteriors as 
$\hat{q} \doteq \{\hat{q}(f^\ell)\}_{\ell=1}^{L}$.

The ELBO for the deep GP under doubly stochastic variational inference can be computed analogously to the single-layer case \eqref{eqn:elbo}.
This work focuses on multifidelity modelling, so we evaluate the ELBO over all
layers~\citep{cutajar2019deep},
giving
\begin{align}\label{eqn:dselbo}
\mathcal{L_{\mathrm{DGP-ELBO}}}
    \left(q(f,u)\right) &= \E_{q(\{f^\ell,U^\ell\}_{\ell=1}^L)}\log\left(\frac{p\left(Y,\{f^\ell,U^\ell\}_{\ell=1}^L\right)}{q\left(\{f^\ell,U^\ell\}_{\ell=1}^L\right)}\right)
\\ \nonumber
    &= \sum_{\ell=1}^{L}\sum_{i=1}^{n^\ell}\E_{q(f^\ell_i)}\log p(Y_i\gvn f^\ell_i) - \beta \sum_{\ell=1}^L\mathcal{D_{\mathrm{KL}}}\left(\phi(U^\ell)\,\|\,p(U^\ell)\right),
\end{align}
where a $\beta$ scaling term has been introduced to modulate the regularizing effect of the KL divergence term.
The expectations in \eqref{eqn:dselbo} can be approximated using the samples over all layers
$\hat{f}$.

More recently, an alternative training objective that restores full symmetry between the objective itself and the predictive posterior has been introduced \citep{jankowiak2020parametric}. 
This objective, known as the predictive log likelihood (PLL), is obtained by directly minimizing the KL divergence from the empirical output distribution to the predictive
distribution, then adding the KL divergence regularizing term from the ELBO~\citep{jankowiak2020parametric}:
\begin{align}
\label{eq:pll}
\mathcal{L}_\mathrm{DGP-PLL}
    &= \mathbb{E}_{p_{\text{data}}\left(\{Y^l\}_{\ell=1}^L,\, X\right)}
    \left( \log q(\{Y^\ell\}_{\ell=1}^L \gvn X) \right)
    - \beta \sum_{\ell=1}^L \mathcal{D}_{\mathrm{KL}}\left(q(U^\ell) \parallel p(U^\ell)\right),
\\ \nonumber
    &= \sum_{\ell=1}^L \sum_{i=1}^{n_\ell}
        \left( \log q(Y^\ell \gvn X_i) \right)
    - \beta \sum_{\ell=1}^L \mathcal{D}_\mathrm{KL}\left(q(U^\ell) \parallel p(U^\ell)\right).
\end{align}
This objective gives good performance for single-layer GPs and can also be applied to deep GPs \citep{jankowiak2020deep}.
However, for deep GPs, the expectation over the latent function values occurs inside the log, so approximation of $\mathcal{L}_\mathrm{DGP-PLL}$ using finite samples from $q(f^{l-1} \gvn x)$ results in a biased estimator.
Recent work developed the deep sigma point process to address this issue \citep{jankowiak2020deep}, but here we found that approximating the expectations with sufficiently many samples gives adequate performance.


\section{Gradient-enhanced deep GP}
\label{sec:gradient_enhanced_deep_gp}

In view of the methods introduced in
Sec.\,~\ref{sec:gaussian_process_regression}, we construct our method for
extending deep GPs to incorporate gradient data.
Deep GPs can be readily extended to incorporate gradient information by
predicting and conditioning on $f^\ell_\nabla$ each layer, where
\begin{align}\label{eq:f_grad}
f^\ell_{\nabla} = [f^\ell,\nabla^\top f^\ell]^\top \in \R^{d+1}
\end{align}
and $\nabla f^\ell$ is the gradient of $f^\ell$ with respect to the base inputs $x$.
Gradients can be predicted by modifying both the deep GP kernel function and
variational posterior and passing the function values and gradients
\eqref{eq:f_grad} through each layer.
The first layer takes only the base inputs $x$ and thus uses a standard
gradient-enhanced kernel \eqref{eq:grad_k}.
For the subsequent layers $\ell\geq2$, the corresponding gradient kernels
$k^\ell_{\nabla} : \R^{d+1} \times \R^d \times \R^{d+1} \times \R^d \to \R^{(d+1)\times(d+1)}$ satisfy
\begin{equation}
\label{eq:dgp_kl_grad}
k^\ell_{\nabla}\big((f^{\ell-1}_{\nabla p},\, x_p),\, (f^{\ell-1}_{\nabla q},\, x_q)\big) =
    \begin{bmatrix}
    k^\ell\big((f^{\ell-1}_p,\, x_p),\, (f^{\ell-1}_q,\, x_q)\big) & \nabla_{q}^\top k^\ell\big((f^{\ell-1}_p,\, x_p),\, (f^{\ell-1}_q,\, x_q)\big)
    \\
    \nabla_{p} k^\ell\big((f^{\ell-1}_p,\, x_p),\, (f^{\ell-1}_q,\, x_q)\big) &
        \nabla_{p} \nabla_{q}^\top k^\ell\big((f^{\ell-1}_p,\, x_p),\, (f^{\ell-1}_q,\, x_q)\big)
    \end{bmatrix}.
\end{equation}
Due to the separable structure of the kernel \eqref{eq:dgp_kernel_form},
the gradient terms can be explicitly computed as
\begin{align}
\nabla_{p} k^\ell\big((f^{\ell-1}_{\nabla p},\, x_p),\, (f^{\ell-1}_{\nabla q},\, x_q)\big) &=
    \nabla_{p} \left(k^\ell_{g x}(x_p,\, x_q) \cdot k^\ell_{g f}(f^{\ell-1}_p,\, f^{\ell-1}_q)\right)
    + \nabla_{p}\, k^\ell_{\gamma x}(x_p,\, x_q)
\\ \nonumber
    &= \nabla_{p} k^\ell_{g x}(x_p,\, x_q) \cdot k^\ell_{g f}(f^{\ell-1}_p,\, f^{\ell-1}_q)
        + k^\ell_{g x}(x_p,\, x_q) \cdot \nabla_{p} k^\ell_{g f}(f^{\ell-1}_p,\, f^{\ell-1}_q)
\\ \nonumber
&\quad + \nabla_{p}\, k^\ell_{\gamma x}(x_p,\, x_q).
\end{align}
Explicitly accounting for the functional dependence of $f_p^{\ell-1}$ on $x_p$
--- i.e.,\, $f_p^{\ell-1} = f^{\ell-1}(x_p)$ --- using the chain rule gives
\begin{equation}
\label{eq:grad_x_k_rho}
\nabla_{p} k^\ell_{g f}(f^{\ell-1}_p,\, f^{\ell-1}_q) = \nabla_{f^{\ell-1}_p} k^\ell_{g f}(f^{\ell-1}_p,\, f^{\ell-1}_q) \cdot \nabla f^{\ell-1}(x_p).
\end{equation}
By denoting gradient kernels with respect to their inputs as 
$k^\ell_{gx \nabla}$, $k^\ell_{gf \nabla}$, and $k^\ell_{\gamma
\nabla}$ and defining matrices
\begin{equation}
\hat{F}_{\nabla p}^\ell =
\begin{bmatrix}
    1 & 0 \\ \mbf{0} & \nabla f^{\ell-1}(x_p)
    \end{bmatrix}
    \qquad \text{and} \qquad
\hat{F}_{\nabla q}^\ell =
\begin{bmatrix}
    1 & 0 \\ \mbf{0} & \nabla f^{\ell-1}(x_q)
    \end{bmatrix},
\end{equation}
\eqref{eq:dgp_kl_grad} becomes 
\begin{align}
\label{eq:dgp_kl_grad_clean}
k^\ell_{\nabla}\big((f^{\ell-1}_{\nabla p},\, x_p),\, (f^{\ell-1}_{\nabla q},\, x_q)\big) &=
    k^\ell_{g x \nabla}(x_p,\, x_q)  \cdot k^\ell_{g f}(f^{\ell-1}_p,\, f^{\ell-1}_q)
\\ \nonumber
    &\quad + k^\ell_{g x}(x_p,\, x_q)  \cdot 
    \hat{F}_{\nabla p}^\ell\,
    k^\ell_{g f \nabla}(f^{\ell-1}_p,\, f^{\ell-1}_q)\,
    (\hat{F}_{\nabla q}^\ell)^\top
\\ \nonumber
    &\quad + k^\ell_{\gamma x \nabla}(x_p,\, x_q).
\end{align}
The choice of affine mean used in the deep GP \eqref{eq:affine_mean}
means the gradient-enhanced mean function $m_{\nabla}^l : \R^{d+1} \times \R^d \to \R^{d+1}$
is simply
\begin{equation}
\label{eq:affine_mean_grad}
m_{\nabla}^\ell(f^{\ell-1}_{\nabla}(x),\, x) =
\begin{bmatrix}
\kappa f^{\ell-1}(x) + c
\\
\kappa \nabla f^{\ell-1}(x)
\end{bmatrix}.
\end{equation}

We now define a variational posterior where each inducing input is augmented
with additional `inducing gradients'.
For each base-layer inducing input $z^1_i$, we define the
corresponding inducing input $(z_{\nabla}^\ell)_i \in \R^{d+1}$ for each layer
$\ell \geq 2$ and write
$Z^\ell_\nabla = \big\{(z_{\nabla}^\ell)_i\big\}_{i=1}^{m_\ell}$.
When using a full-rank variational posterior, we fix the inducing points to the
training data locations and set these additional $d$ inducing values for each
gradient data point, so we abuse notation and write 
$z_i^{\ell} = [Y_i^{\ell-1}, \nabla^\top Y_i^{\ell-1}]^\top$.
Analogously to \eqref{eq:variational_posterior}, we set the form of the gradient-enhanced variational posterior as
\begin{align}
q\Big(\big\{{f}_\nabla^\ell,\, {U}_\nabla^\ell \big\}_{\ell=1}^L\Big) =
    \prod_{\ell=1}^L p\big({f}_\nabla^\ell \gvn {U}_\nabla^\ell\big)\,
q\big({U}_\nabla^\ell\big),
\end{align}
where
$q\big({U}_\nabla^\ell\big) = \mathcal{N}\big({m}_{\nabla q}^\ell,\,
S_{\nabla q}^\ell\big)$,
with mean ${m}_{\nabla q}^\ell \in \R^{m^\ell(d+1)}$ and covariance
$S_{\nabla q}^\ell \in \R^{m_\ell(d+1) \times m_\ell(d+1)}$.
As these terms factorize layer-wise and all distributions are Gaussian, the inducing values can be marginalized out to obtain
\begin{equation}
\label{eq:deep_grad_gp_f}
q\Big(\big\{ {f}_\nabla^\ell \}_{\ell=1}^L\Big)
    = \prod_{\ell=1}^L q\big({f}_\nabla^\ell\big) 
    = \prod_{l=1}^L \mathcal{N}\big(\tilde{\mu}_\nabla^\ell,\,
\tilde{\Sigma}_\nabla^\ell\big).
\end{equation}
Analogously to the single-layer case, the mean and covariance terms are
\begin{align}
\label{eq:q_grad_fl_marginal}
(\tilde{\mu}^\ell_\nabla)_{i} &= m_{\nabla}^\ell(f^{\ell-1}_{\nabla i},\, X_i) + \alpha_\nabla(f^{\ell-1}_{\nabla i},\, X_i)^\top ({m}^\ell_{\nabla q} - m_{\nabla}^\ell(z^\ell_{\nabla i},\, z^1_i))
\\ \nonumber
(\tilde{\Sigma}^l_\nabla)_{ij}
    &= k^\ell_{\nabla}((f^{\ell-1}_{\nabla i},\, X_i),\, (f^{\ell-1}_{\nabla j},\, X_j)) - \alpha_\nabla(f^{\ell-1}_{\nabla i},\, X_i)^\top
(K_{Z_\nabla^\ell Z_\nabla^\ell} - S^\ell_{\nabla q})\, 
\alpha_\nabla(f^{\ell-1}_{\nabla j},\, X_j),
\end{align}
where $\alpha_\nabla(f^{\ell-1}_{\nabla i},\, X_i) = K_{Z_\nabla^\ell Z_\nabla^\ell}^{-1} {K}_{Z_\nabla^\ell f^{l-1}_{\nabla i}}$ and the matrices $K_{Z^l_\nabla Z^l_\nabla }$ and 
${K}_{Z^\ell_\nabla f^{l-1}_{\nabla i}}$ have entries
\begin{align*}
(K_{Z^l_\nabla Z^l_\nabla })_{pq} &= k^\ell_{\nabla}\big((z^\ell_{\nabla p},\, z^1_p),\, (z^l_{\nabla q},\, z^1_q))
\\ \nonumber
\big({K}_{Z^\ell_\nabla f^{l-1}_{\nabla i}})_{pq} &= k^\ell_{\nabla}\big((f^{l-1}_{\nabla p},\, X_p),\, (z^l_{\nabla q},\, z^1_q)\big).
\end{align*}

Under this construction, $f_{\nabla i}^\ell$ depends only on
$f_{\nabla i}^k$ for $k <\ell$,
so we can sample from $q\big(f_{\nabla i}^l\big)$ by iterative computation from the base layer to
layer $\ell$, using samples from the marginal distributions
$q\big(f_{\nabla i}^k\big)$ as inputs to the next layer.
We remark that for most choices of base kernel, and in particular the
squared-exponential, the off-diagonal terms in
$k^\ell_{\nabla}\big((f^{\ell-1}_{\nabla i},\, x_i),\, (f^{\ell-1}_{\nabla i},\, x_i)\big)$
vanish, enabling independent sampling from each $q\big(f_{\nabla i}^\ell\big)$ and thus mitigating
scaling issues associated with high input dimensions $d$.
Under our choice of variational distribution, the multifidelity gradient-enhanced ELBO is 
\begin{equation}
\label{eq:mf_dgp_grad_lower_bound}
\mathcal{L}_{\nabla DGP-ELBO}\Big(q\big(\{{f}_\nabla^\ell,\, {U}_\nabla^\ell \}_{\ell=1}^L\big)\Big) =
    \sum_{\ell=1}^L \sum_{i=1}^{n_\ell} \mathbb{E}_{q\big(f_{\nabla i}^\ell\big)}
\log p\big(y_{\nabla i}^\ell \gvn f_{\nabla i}^\ell\big)
    - \sum_{\ell=1}^L \mathcal{D}_{\mathrm{KL}}\Big(q\big({U}_\nabla^\ell\big) \big\|
p\big({U}_\nabla^\ell\big)\Big),
\end{equation}
where expectations over each $q(f^\ell_{\nabla i})$ are approximated by sampling from the
previous layer's distribution $q(f^{\ell-1}_{\nabla i})$.
The gradient-enhanced PLL objective is defined from \eqref{eq:pll} analogously.


\section{Implementation}
\label{sec:implementation}

This section briefly summarizes the implementation of the multifidelity GPR
models presented in Sec.\;\ref{sec:gaussian_process_regression} and
\ref{sec:gradient_enhanced_deep_gp}.
For fidelities $\ell = 1,...,L$, we compute the output data
$(Y^\ell_i)_{i=1}^{n^\ell}$
by sampling the model $\tilde{f}^\ell$ at the chosen input locations
$(X^\ell_i)_{i=1}^{n^\ell}$.
We denote the input and output data over all layers as
$\mathcal{X} = \{(\ell, X^\ell)\}_{\ell=1}^L$ and 
$\mathcal{Y} = \{Y^\ell\}_{\ell=1}^L$.
Moreover, we denote the desired prediction locations paired with the highest
fidelity level as $\mathcal{X}^\star = (L, X^\star)$.
Initial values for the hyperparameters and variational parameters are denoted
$\theta_0$ and $\psi_0$ respectively.

Alg.\;\ref{alg:lmc}, \texttt{ComputeLMCPredictions},
describes how the LMC models are constructed from the data.
\texttt{ComputeKLMC($\mathcal{X}_1$, $\mathcal{X}_2$, $\theta$)}
computes the matrix $K_{LMC}$ for the fidelity-input pairs $\mathcal{X}_1$,
$\mathcal{X}_2$ and hyperparameters $\theta$ using the LMC kernel function
\eqref{eq:lmc_cov};
\texttt{ComputeMLL($K$, $Y$, $\theta$)} computes the marginal log likelihood for the
gram matrix $K$, output data $Y$, and hyperparameters $\theta$ using
\eqref{eq:log_ml}; and 
\texttt{ComputeFStar($K$, $K^\star$, $K^{\star\star}$, $Y$)} computes the marginal
predictive mean and covariance for gram matrices
$K$, $K^\star$, $K^{\star\star}$ and output data $Y$ using
\eqref{eq:posterior_mk};
\texttt{OptimizerUpdate}($L$, $\theta$) updates the parameters $\theta$ using a
gradient-based optimizer to reduce the loss $L$.
For a gradient-enhanced LMC model, 
the gradient kernel $k_\nabla$ \eqref{eq:grad_k} is used in place of a standard kernel when
evaluating \texttt{ComputeKLMC} and gradient data is included in output data
$\mathcal{Y}$.

Alg.\;\ref{alg:dgp_posterior}, \texttt{ComputeDeepGPPosterior}, shows how to
compute the approximate deep GP
variational posterior from input points $X$, hyperparameters $\theta$,
variational parameters $\psi$, and number of samples $N_S$.
\texttt{ComputeMDGP}($X$, $f$, $\theta$) evaluates the deep GP mean function
\eqref{eq:affine_mean} for base inputs $X$, previous-layer outputs $f$, and
hyperparameters $\theta$;
\texttt{ComputeKDGP}($X_1$, $X_2$, $f_1$, $f_2$, $\theta$) evaluates the deep GP
kernel function \eqref{eq:dgp_kernel_form} for base inputs $X_1$ and $X_2$,
previous-layer outputs $f_1$ and $f_2$, and hyperparameters $\theta$;
\texttt{ComputeQ}($m_{Z}$, $m_{X}$, $K_{ZZ}$, $K_{ZX}$, $\psi$)
computes the variational posterior \eqref{eq:dgp_mu_k}
for inducing mean $m_{Z}$, mean $m_{X}$, gram matrices $K_{ZZ}$ and $K_{ZX}$
hyperparameters $\psi$;
\texttt{SampleQ}($q$) samples the multivariate distribution $q$ by sampling from
each variable independently \eqref{eq:q_sampling};
\texttt{ComputeMixture}($q_1, ..., q_N$) computes the mixture model
from the input distributions according to \eqref{eq:approx_q}.

Alg.\;\ref{alg:dgp} describes how the deep GP models are built from the data.
\texttt{EvaluateDGPObjective}($\mathcal{L}$, $\hat{f}$, $\mathcal{Y}$) evaluates
the chosen objective function $\mathcal{L}$ with posterior samples $\hat{f}$ and
data $\mathcal{Y}$.
$\mathcal{L}$ is chosen as either the ELBO \eqref{eqn:dselbo} or PLL objective
\eqref{eq:pll}.
To build a gradient-enhanced deep GP model, 
the gradient deep GP kernel $k_\nabla^\ell$ \eqref{eq:dgp_kl_grad}
and mean $m_{\nabla}^\ell$ \eqref{eq:affine_mean_grad}
are used when evaluating \texttt{ComputeKDGP} and \texttt{ComputeMDGP},
the gradient-enhanced variational posterior is used when evaluating
\texttt{ComputeQ} \eqref{eq:q_grad_fl_marginal},
gradient data is included in the output data $\mathcal{Y}$,
and a gradient objective function is used (see
\eqref{eq:mf_dgp_grad_lower_bound} for the ELBO).

The models are implemented using
\texttt{gpytorch}~\citep{gardner2018gpytorch}, which is built on the
\texttt{pytorch}~\citep{paszke2019pytorch} framework.
Training is performed 
using the Adam
optimizer~\citep{kingma2014adam}, with the required derivatives computed using
\texttt{pytorch}'s automatic differentiation tools.
We found that a four-stage training procedure on normalized data was typically
sufficient to converge the optimization problems.
These stages consisted of $N_T=800$ iterations with learning rates of 0.03, 0.01,
0.003, and 0.001 respectively.

\begin{algorithm}
\DontPrintSemicolon
\small
\KwData{$\mathcal{X}^\star$, $\mathcal{X}$, $\mathcal{Y}$, $N_T$, $\theta_0$}
\KwResult{$f^L(X^\star)\,|\, \mathcal{X},\mathcal{Y}$
\tcp{high-fidelity predictive distribution}}
$\theta \gets \theta_0$ \tcp{initialize hyperparameters}
\For{$t$ in $1,\, ...,\, N_T$}{
$K_{LMC} \gets \texttt{ComputeKLMC}(\mathcal{X}, \mathcal{X}, \theta)$
    \tcp{evaluate gram matrix}
$L \gets \texttt{ComputeMLL}(K_{LMC}, \mathcal{Y}, \theta)$
    \tcp{evaluate log-marginal likelihood}
$\theta \leftarrow \texttt{OptimizerUpdate}(L, \theta)$
    \tcp{update hyperparameters}
}
$K_{LMC} \gets \texttt{ComputeKLMC}(\mathcal{X}, \mathcal{X}, \theta)$,
$K_{LMC}^\star \gets \texttt{ComputeKLMC}(\mathcal{X}^\star, \mathcal{X}, \theta)$,
$K_{LMC}^{\star\star} \gets \texttt{ComputeKLMC}(\mathcal{X}^\star, \mathcal{X}^\star, \theta)$\;
$\mu_{f^\star}, \Sigma_{f^\star} \gets \texttt{ComputeFStar}(K_{LMC}, K_{LMC}^\star,
K_{LMC}^{\star\star}, \mathcal{Y})$
\tcp{posterior mean and covariance}
$f^L(X^\star)\,|\, \mathcal{X},\mathcal{Y} \gets \N(\mu_{f^\star}, \Sigma_{f^\star})$\;
\Return{$f^L(X^\star)\,|\, \mathcal{X},\mathcal{Y}$ \tcp{predictive distribution}}
\caption{\texttt{ComputeLMCPredictions}\label{alg:lmc}
}
\end{algorithm}

\begin{algorithm}
\DontPrintSemicolon
\small
\KwData{$X$, $\theta$, $\psi$, $N_S$}
\KwResult{
$\hat{q}$, $\hat{f}$ 
\tcp{variational posterior and corresponding function samples}
}
\tcp{Compute first-layer output distribution}
$m^1_{Z} \gets \texttt{ComputeMDGP}(Z^1, \texttt{None}, \theta)$\;
$m^1_{X} \gets \texttt{ComputeMDGP}(X,   \texttt{None}, \theta)$\;
$K^1_{ZZ} \gets \texttt{ComputeKDGP}(Z^1, Z^1, \texttt{None}, \texttt{None}, \theta)$\;
$K^1_{ZX} \gets \texttt{ComputeKDGP}(Z^1, X,   \texttt{None}, \texttt{None}, \theta)$\;
$q(f^1) \gets \texttt{ComputeQ}(m^1_{Z}, m^1_{X}, K^1_{ZZ}, K^1_{ZX}, \psi)$\;
\tcp{Compute $N_S$ posterior samples through all layers}
\For{$\ell$ in $2,\, ...,\, L$}{
    $m^\ell_{Z} \gets \texttt{ComputeMDGP}(Z^1, Z^\ell, \theta)$\;
    $K^\ell_{ZZ} \gets \texttt{ComputeKDGP}(Z^1, Z^1, Z^\ell, Z^\ell, \theta)$\;
    \For{$j$ in $1,\, ...,\, N_S$}{
    $(\hat{f}^1)^j \gets \texttt{SampleQ}(q(f^1))$\;
        $m^\ell_{X} \gets \texttt{ComputeMDGP}(X,   (\hat{f}^{\ell-1})^j, \theta)$\;
        $K^\ell_{ZX} \gets \texttt{ComputeKDGP}(Z^1, X, Z^\ell, (\hat{f}^{\ell-1})^j, \theta)$\;
        $\hat{q}^j(f^\ell) \gets \texttt{ComputeQ}(m^\ell_{Z}, m^\ell_{X}, K^\ell_{ZZ}, K^\ell_{ZX}, \psi)$\;
        $(\hat{f}^\ell)^j \gets \texttt{SampleQ}(\hat{q}^j(f^\ell))$\;
    }
}
$\hat{q} \gets \texttt{ComputeMixture}(\hat{q}^1(f^1), ..., \hat{q}^{N_S}(f^L))$
  \tcp{make approximate posterior}
\Return{$\hat{q}$, $\hat{f}$}
\caption{\texttt{ComputeDeepGPPosterior}
\label{alg:dgp_posterior}}
\end{algorithm}

\begin{algorithm}
\DontPrintSemicolon
\small
\KwData{$\mathcal{X}^\star$, $\mathcal{X}$, $\mathcal{Y}$, $N_T$, $N_S$,
$N^\star_S$, $\theta_0$, $\psi_0$, $\mathcal{L}$}
\KwResult{
$f^L(X^\star)\,|\, \mathcal{X},\mathcal{Y}$
\tcp{high-fidelity predictive distribution}
}
$\theta \gets \theta_0$, $\psi \gets \psi_0$
  \tcp{initialize hyperparameters and variational parameters}
$X_0 \gets \bigcup \mathcal{X}$ \tcp{evaluate DGP at all points}
\For{$t$ in $1,\, ...,\, N_T$}{
    $\hat{q}, \hat{f} \gets \texttt{ComputeDGPPosterior}(X_0, \theta, \psi, N_S)$
      \tcp{compute approximate posterior}
    $L \gets \texttt{EvaluateDGPObjective}(\mathcal{L}, \hat{f}, \mathcal{Y})$
      \tcp{evaluate objective using samples}
    $\theta, \psi \leftarrow \texttt{OptimizerUpdate}(L, (\theta, \psi))$
        \tcp{update hyperparameters and variational parameters}
}
$X^\star_0 \gets \bigcup \mathcal{X}^\star$\;
$\hat{q}, \hat{f} \gets \texttt{ComputeDGPPosterior}(X_0^\star, \theta, \psi, N_S^\star)$
    \tcp{predictive distribution}
$f^L(X^\star)\,|\, \mathcal{X},\mathcal{Y} \gets \hat{q}^L$\;
\Return{$f^L(X^\star)\,|\, \mathcal{X},\mathcal{Y}$
\tcp{approximate predictive distribution}
}
\caption{\texttt{ComputeDeepGPPredictions}\label{alg:dgp}}
\end{algorithm}

\clearpage
\section{Numerical examples}
\label{sec:results}

\subsection{Results: Test problem}
\label{sec:results_test_problem}

\begin{figure}[h]
\begin{minipage}{0.57\textwidth}
\centering
\includegraphics[width=0.88\textwidth]{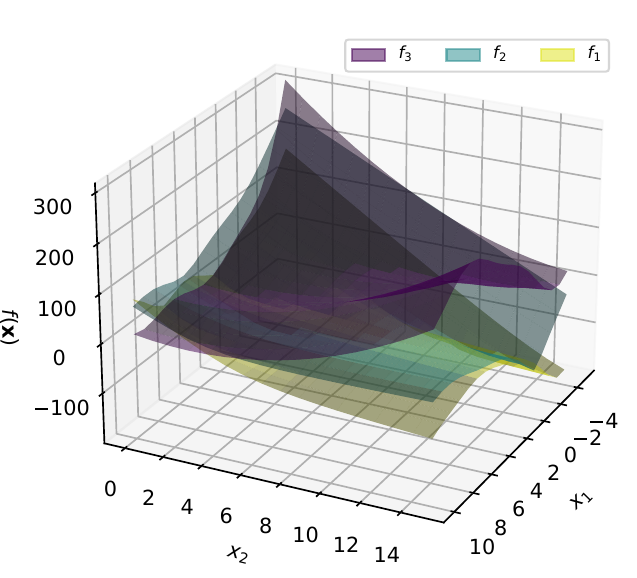}
\vspace{0mm}
\caption{Multifidelity Branin function}
\label{fig:mf_branin}
\end{minipage}
\hfill
\begin{minipage}{0.42\textwidth}
\centering
\includegraphics[width=0.80\textwidth]{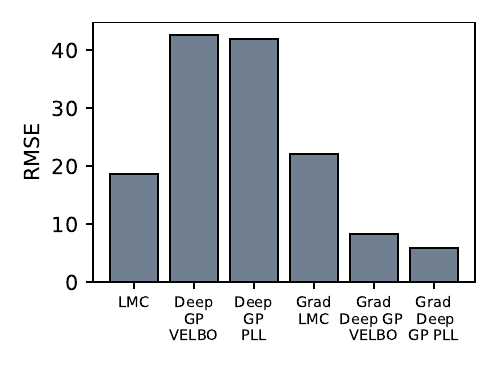}
\vspace{-2mm}
\caption{Prediction errors on multifidelity Branin
function\label{tab:test_problem_results}, medium sampling
}
\label{fig:branin_rmse_plot}
\end{minipage}
\end{figure}

We begin by testing our method on a benchmark test problem, the multifidelity
Branin function (Eq.\;\ref{eq:mf_branin})~\citep{perdikaris2017nonlinear}, to
highlight the differences between each of the GP models. 
The high-fidelity data is sampled from the Branin function and
the medium- and low-fidelity data are generated by nonlinearly transforming the
high-fidelity function.
The multifidelity Branin function is shown in Fig.\;\ref{fig:mf_branin};
note how the relationships between the different fidelities vary over the input
space.
\begin{align}
\label{eq:mf_branin}
f_3(x) & = \left( \frac{-1.275 x_1^2}{\pi^2} + \frac{5 x_1 }{\pi} + x_2 - 6 \right)^2
    + \left( 10 - \frac{5}{4 \pi} \right)\cos(x_1) + 10
\\ \nonumber
f_2(x) &= 10 \sqrt{f_3(x-2)} + 2(x_1 - 0.5) - 3(3 x_2 - 1) - 1
\\ \nonumber
f_1(x) &= f_2(1.2(x+2)) - 3 x_2 + 1
\\ \nonumber
x &= [x_1,\, x_2]^\intercal, \quad -5 \leq x_1 \leq 10, \quad 0 \leq x_2 \leq 15
\end{align}

\vspace{-5mm}

We investigate the predictive performance of the GP models for
three sampling densities --- dense, medium, and sparse --- which are detailed in 
Table\;\ref{tab:mf_branin_results}.
Here, $N_{f_1}$, $N_{f_2}$, and $N_{f_3}$ denote the number of sample points for each function in \eqref{eq:mf_branin}.
For the medium and sparse datasets, we use a full-rank variational posterior
with the number of inducing points equal to the number of data points.
For the dense dataset, we set $m^l = 40$ and use the inducing points as free
variational parameters.
We evaluate the corner points and centre point of the domain, then add the
additional points by uniformly sampling the input space.
The predictions for the LMC and deep GP models are generated using
Alg.\;\ref{alg:lmc} and Alg.\;\ref{alg:dgp} respectively.
See Sec.\;\ref{sec:implementation} for implementations details pertaining to the
gradient-enhanced models.

The table gives root mean square prediction error (RMSE) and mean
absolute prediction error (MAE), evaluated on 100 unseen randomly
distributed test points for $f_3$.

The LMC models use three separable kernels ($R=3$) and thus these
models are generalizations of AR1.
We also compare performance for both choices of deep GP training objective: the
variational ELBO and the PLL.
We set $\beta = 1$ for both choices and use 30 Monte Carlo samples to
approximate the expectations.
For all cases, the gradient-enhanced deep GP methods outperform the LMC
analogue,
while the standard LMC model outperforms both the standard deep GP models.
Fig.\;\ref{fig:branin_rmse_plot} summarizes the performance of the models in terms of RMSE for
the `medium' sampling density.
The choice of deep GP objective function does not significantly affect the mean
prediction errors of the deep GP models for most cases, although the
gradient-enhanced model trained with the PLL objective performance sufffers in the sparse sampling case.

\begin{table}[h]
\renewcommand{\arraystretch}{0.8}
\linespread{1.0}\selectfont\centering
{\footnotesize \caption{Multifidelity Branin
results~\label{tab:mf_branin_results}}
\vspace{0mm}
\begin{tabular}{llllrr}
\toprule
\textbf{Sampling} & $N_{f_1}$, $N_{f_2}$, $N_{f_3}$ & \textbf{Model} & \textbf{Objective} & \textbf{RMSE} & \textbf{MAE} \\
\midrule                                          

Sparse    & 20, 10, 5  & LMC       &            & 27.956          & 20.756  \\
          &            & LMC grad  &            & 19.359          & 15.950  \\
          &            & DGP       & ELBO      & 58.843          & 51.018    \\
          &            & DGP grad  & ELBO      & \textbf{11.236} & \textbf{8.396} \\
          &            & DGP       & PLL        & 48.712          & 40.139    \\
          &            & DGP grad  & PLL        & 20.671          & 15.450    \\
\midrule                                         
Medium    & 40, 20, 10 & LMC       &            & 18.704          & 9.488  \\
          &            & LMC grad  &            & 22.131          & 6.681  \\
          &            & DGP       & ELBO      & 41.471          & 24.100       \\
          &            & DGP grad  & ELBO      & 8.287           & \textbf{2.366}   \\
          &            & DGP       & PLL        & 41.873          & 23.425          \\
          &            & DGP grad  & PLL        & \textbf{5.956}  & 2.475 \\
\midrule                                         
Dense     & 80, 40, 20 & LMC       &            & 7.196           & 4.576  \\
          &            & LMC grad  &            & 6.749           & 2.753  \\
          &            & DGP       & ELBO      & 28.395          & 18.808      \\
          &            & DGP grad  & ELBO      & \textbf{2.377}  & \textbf{1.049} \\
          &            & DGP       & PLL        & 8.652           & 4.722           \\
          &            & DGP grad  & PLL        & 3.244           & 1.081          \\
\bottomrule
\end{tabular}
}
\end{table}

\begin{figure}[h]
\centering
    \begin{adjustbox}{minipage=\linewidth,scale=1.0}
    \begin{subfigure}[t]{0.49\textwidth}
    \centering
    \includegraphics[width=\textwidth]{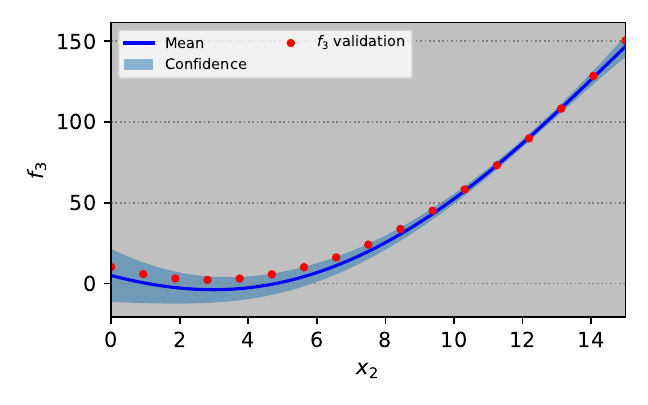}
    \vspace{-7mm}
    \caption{Gradient LMC GP [$x_1=2.5$]}
    \end{subfigure}%
    \hfill
    \begin{subfigure}[t]{0.49\textwidth}
    \centering
    \includegraphics[width=\textwidth]{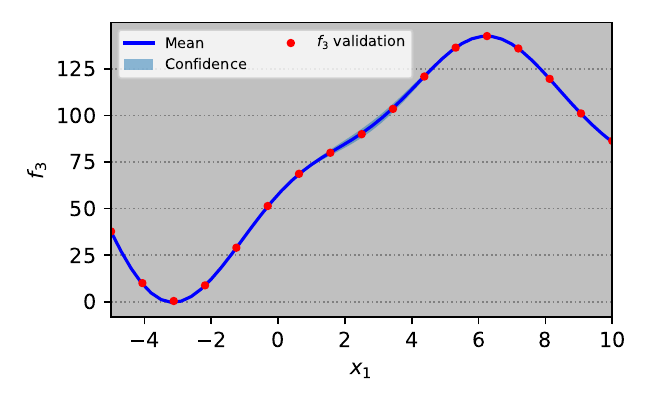}
    \vspace{-7mm}
    \caption{Gradient LMC GP [$x_2=12$]}
    \end{subfigure}
    \vfill
    \begin{subfigure}[t]{0.49\textwidth}
    \centering
    \includegraphics[width=\textwidth]{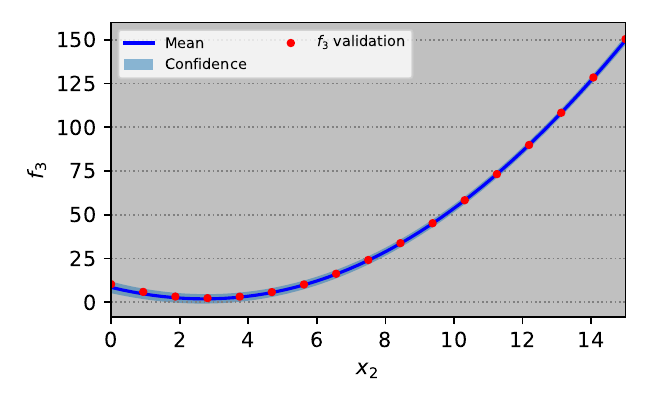}
    \vspace{-7mm}
    \caption{Gradient deep GP, ELBO [$x_1=2.5$]}
    \end{subfigure}%
    \hfill
    \begin{subfigure}[t]{0.49\textwidth}
    \centering
    \includegraphics[width=\textwidth]{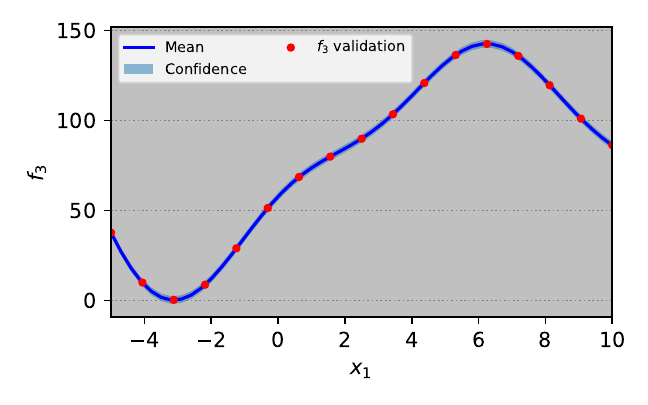}
    \vspace{-7mm}
    \caption{Gradient deep GP, ELBO [$x_2=12$]}
    \end{subfigure}
    \vfill
    \begin{subfigure}[t]{0.49\textwidth}
    \centering
    \includegraphics[width=\textwidth]{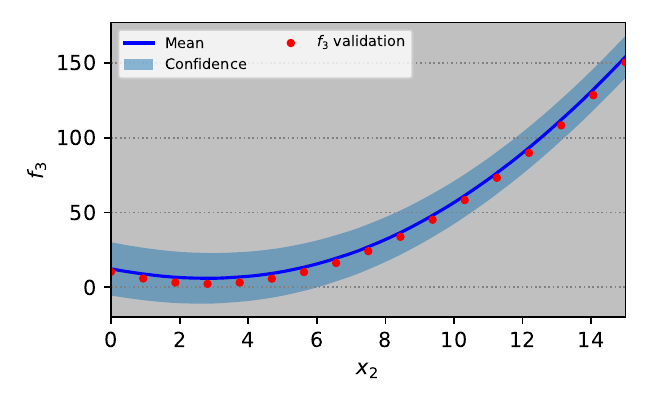}
    \vspace{-7mm}
    \caption{Gradient deep GP, PLL [$x_1=2.5$]}
    \end{subfigure}%
    \hfill
    \begin{subfigure}[t]{0.49\textwidth}
    \centering
    \includegraphics[width=\textwidth]{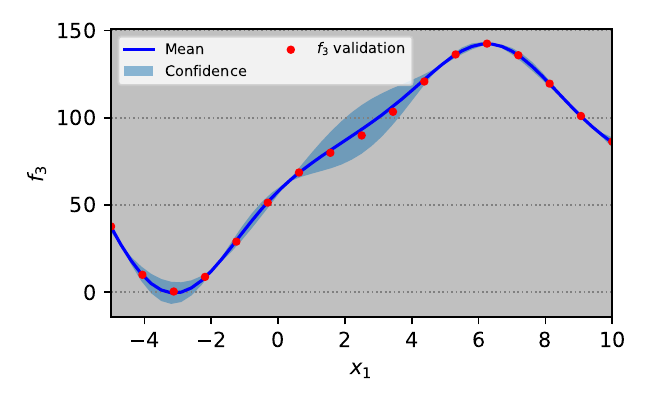}
    \vspace{-7mm}
    \caption{Gradient deep GP, PLL [$x_2=12$]}
    \end{subfigure}
    \end{adjustbox}{}
\vspace{0mm}
\caption{
Performance of gradient-enhanced GP models on the
multifidelity Branin test problem, `medium' sampling density
}
\label{fig:test_problem_slice_comparison}
\end{figure}

To further inspect the differences between the multifidelity techniques,
Fig.\;\ref{fig:test_problem_slice_comparison} shows slices of the
gradient-enhanced models for the `medium' sampling density,
with validation samples from the true function $f_3$ shown in red.
This figure shows the mean predictions and uncertainty bounds, given by the
mean plus or minus two standard deviations, generated by the GP models.
With the ELBO objective, the deep GP model uses output noise to fit the data, resulting in uniform uncertainty bounds.
As discussed in \citep{jankowiak2020parametric}, with the PLL objective, the
model uses internal kernel noise and is better able to account for
input-dependent uncertainty.

\clearpage
\subsection{Results: Aerospace PDE problem}
\label{sec:results_pde_problem}

\begin{figure}[h]
\begin{minipage}{0.44\textwidth}
\centering
\includegraphics[width=0.95\columnwidth]{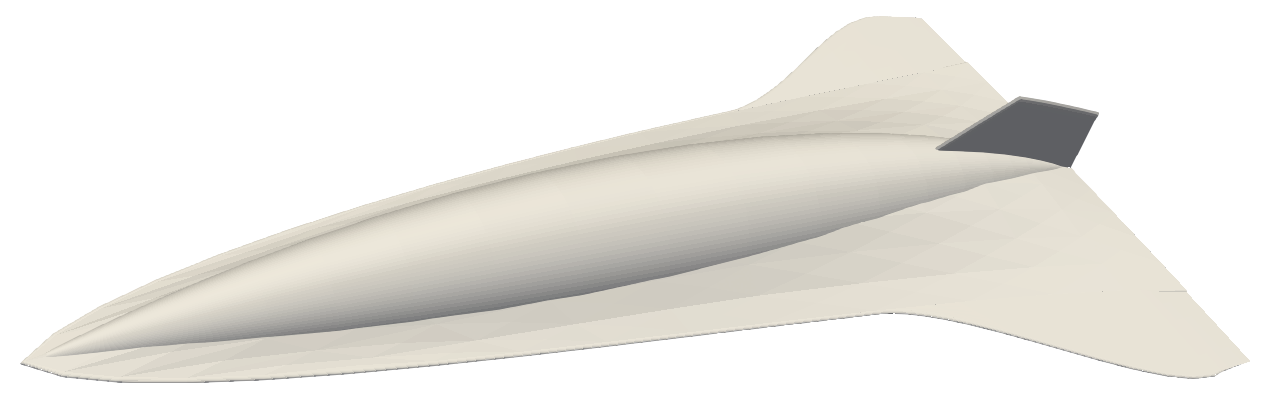}
\caption{Hypersonic waverider geometry}
\label{fig:hypersonic_waverider}
\end{minipage}
\hfill
\begin{minipage}{0.55\textwidth}
\centering
\includegraphics[width=0.95\columnwidth]{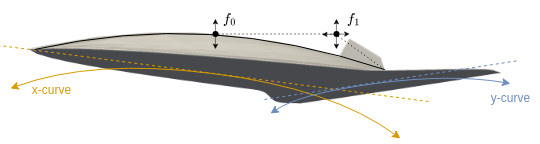}
\caption{Body curvature (exaggerated)}
\label{fig:body_curvature}
\end{minipage}
\end{figure}

We now consider a practical application, where we use various GP models presented herein to approximate
the mapping between boundary conditions for a system of PDEs
and integrated output quantities.
Specifically, we seek to predict the coefficients of lift $C_L$, drag $C_D$, and
pitching moment $C_M$ for a hypersonic waverider vehicle geometry (see
Fig.\;\ref{fig:hypersonic_waverider}) over three inputs: free stream Mach number
Ma, angle of attack AoA, and body curvature $G_c$.
The body curvature input controls the rate of curvature along the
vehicle's longitudinal axis, as illustrated in Fig.\;\ref{fig:body_curvature}.
The training data is generated by evaluating the flow solver at different
levels of mesh discretization,
which naturally leads to approximations of the quantities of interest at different fidelity levels.

The flow field around the vehicle is given by the steady-state solution to the
Euler equations over a domain $\Omega$ enclosing the geometry, closed with the equation
of state for air.
These solutions to the steady-state Euler equations $u:\Omega\to \R^3$ satisfy
\begin{align}
\label{eq:euler_eqs}
\nabla \cdot (\rho u) &= 0
\\ \nonumber
\rho (u \cdot \nabla) u + \nabla p &= 0
\\ \nonumber
\nabla \cdot (\rho u E + p u) &= 0
\end{align}
in $\Omega$, subject to the boundary conditions $u = g(\text{Ma}, \text{AoA}, G_c)$ on $\partial \Omega$.
Here, $\rho:\Omega \to \R$ is the fluid density, $p:\Omega \to \R$ is pressure, $u$ is the velocity
vector, and $E:\Omega \to \R$ is total energy.

The aerodynamic coefficients are computed from the pressure field as
\begin{align}
C_L &= \frac{1}{\frac{1}{2} \rho u_\infty^2 S} \oint (p - p_\infty) n_y \, dS
\\ \nonumber
C_D &= \frac{1}{\frac{1}{2} \rho u_\infty^2 S} \oint (p - p_\infty) n_x \, dS
\\ \nonumber
C_M &= \frac{1}{\frac{1}{2} \rho u_\infty^2 S L} \oint (p - p_\infty) (x - x_{ac}) n_y \, dS
\end{align}
where $p_\infty$ and $U_\infty$ are the free-stream (boundary) pressure and
velocity magnitude, $S$ is the planform area of the vehicle,
$L$ is the vehicle length,
$x_{ac}$ is the coordinate of the centre of pressure, and
$n_x$ and $n_y$ arethe components of the unit normal vector to the surface.

As this example considers supersonic flight conditions, the inflow boundary
is specified by the free-stream air conditions consisting of pressure $p_\infty$ and
temperature $T_\infty$, as well as the inlet Mach number and angle of attack.
We use a zero-gradient outflow boundary condition and a no-penetration wall
boundary condition.

\begin{figure}[h]
\begin{minipage}{0.49\textwidth}
\centering
\includegraphics[width=0.9\textwidth]{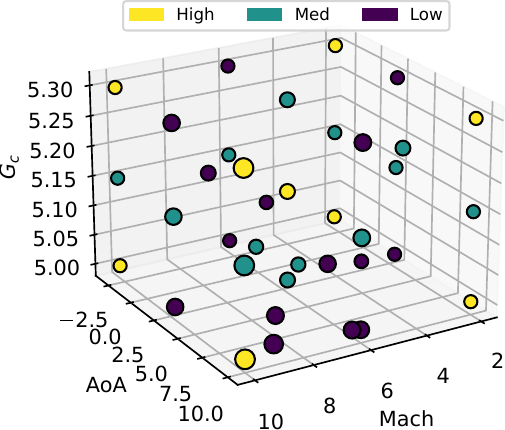}
\caption{Sampling points for aerospace example}
\label{fig:c3d_sampling}
\end{minipage}
\hfill
\begin{minipage}{0.49\textwidth}
\centering
\includegraphics[width=0.90\textwidth]{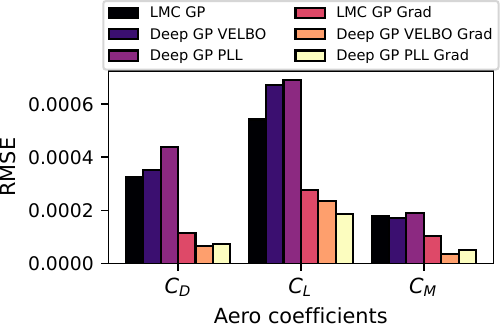}
\caption{RMSE for aerospace example}
\label{fig:rmse_plot}
\end{minipage}
\end{figure}

We use NASA \texttt{CART3D} to compute approximate solutions to \eqref{eq:euler_eqs} and evaluate the aerodynamic coefficients for each set of boundary conditions.
\texttt{CART3D} features adjoint-based adaptive mesh refinement using
cartesian cut cells~\citep{nemec2008adjoint}.
We generate multifidelity data by using this feature to automatically compute
solutions for up to seven mesh refinement levels for each set of inputs.
We employ three fidelity levels: low, medium, and high, which correspond to
three, five, and seven mesh refinements respectively.
Table\;\ref{tab:data_fidelities} lists the nominal runtime for each fidelity
level; the medium and low fidelity data are 5$\times$ and 25$\times$ cheaper
respectively than the high fidelity data.
When using a flow solver without automatic mesh refinement, multifidelity
information may instead be obtained by externally generating several meshes with
different refinement factors, then computing the flow solution for each mesh.
The 37 training points this example, shown in Fig.\;\ref{fig:c3d_sampling}, span
2 $\leq$ Ma $\leq$ 10,
$-3^\circ\leq\text{AoA}\leq10^\circ$, and
5.0 $\leq$ $G_c$ $\leq$ 5.3.
The corner points and centre point are evaluated to high fidelity (seven levels
of mesh refinement) and the remaining points are evaluated to medium or low
fidelity.
The deep GP variants use a full-rank variational posterior, where all data
points used as inducing points.
We use 30 Monte Carlo samples during training and 300 Monte Carlo samples when
evaluating the prediction errors and generating the results plots.
For the gradient-enhanced deep GP, we set the KL divergence scaling factor to
$\beta=2$ for both the ELBO and PLL objectives.
For the standard deep GP, we use the usual setting of $\beta=1$.
Again, the predictions for the LMC and deep GP models are generated using
Alg.\;\ref{alg:lmc} and Alg.\;\ref{alg:dgp} respectively,
with Sec.\;\ref{sec:implementation} detailing implementation of the
gradient-enhanced models.

\begin{table}[h]
\renewcommand{\arraystretch}{0.9}
\linespread{1.0}\selectfont\centering
{\footnotesize \caption{Data fidelities\label{tab:data_fidelities}}
\vspace{0mm}
\begin{tabular}{lrrrr}
\toprule
\textbf{Fidelity} & \textbf{Mesh refinements} & \textbf{Runtime [sec]} & \textbf{Speed increase} \\
\midrule                                          
low    & 3      &  230   &  24.88   \\
medium & 5      &  1230  &  5.005   \\
high   & 7      &  4200  &  1.0      \\
\bottomrule
\end{tabular}
}
\end{table}
Fig.\;\ref{fig:rmse_plot} compares the RMSE prediction error of the
multifidelity GP models and the single-fidelity reference model across the three
outputs --- $C_L$, $C_D$, and $C_M$ --- on a test set of 50 points.
The prediction errors are small, highlighting the effectiveness of GPR models
for this application.
The gradient-enhanced methods outperform the standard methods in all cases,
clearly demonstrating the value of incorporating gradients into GP models.
Moreover, for the gradient-enhanced deep GP, models trained with the PLL objective outperform the
ELBO objective in all cases, corroborating the findings
in~\citep{jankowiak2020parametric}.
Similarly, using the PLL objective, the gradient-enhanced deep GP outperforms the its LMC
analogue, further demonstrating that the deep GP's ability to capture
input-dependent function correlations makes them well-suited for multifidelity modelling.

\begin{figure}[h]
\centering
    \begin{adjustbox}{minipage=\linewidth,scale=1.0}
    \begin{subfigure}[t]{0.32\textwidth}
    \centering
    \includegraphics[width=\textwidth]{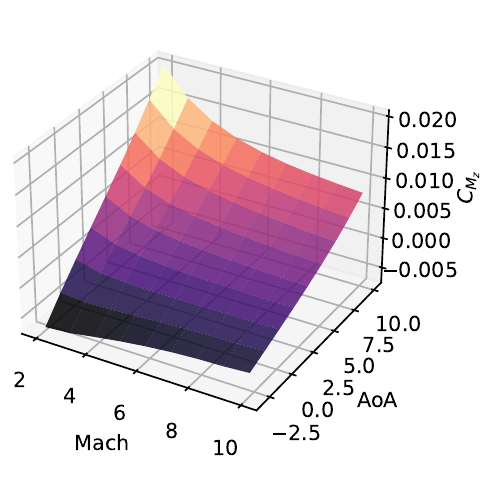}
    \vspace{-6mm}
    \caption{Pitching moment}
    \end{subfigure}%
    \hfill
    \begin{subfigure}[t]{0.32\textwidth}
    \centering
    \includegraphics[width=\textwidth]{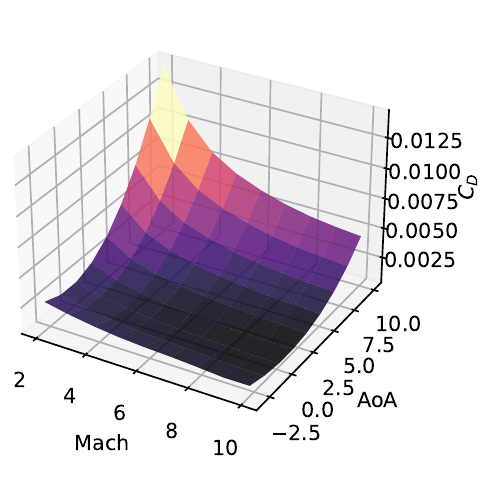}
    \vspace{-6mm}
    \caption{Drag}
    \end{subfigure}
    \hfill
    \begin{subfigure}[t]{0.32\textwidth}
    \centering
    \includegraphics[width=\textwidth]{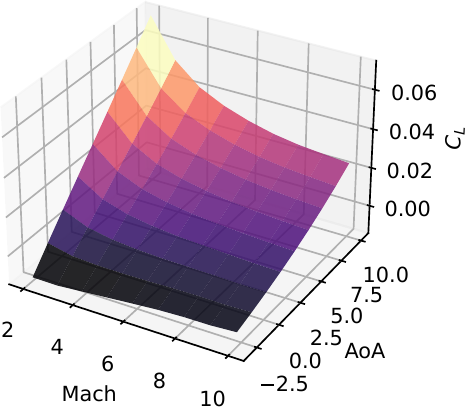}
    \vspace{-6mm}
    \caption{Lift}
    \label{fig:comparison_gpr_reference}
    \end{subfigure}
    \end{adjustbox}{}
\vspace{-1mm}
\caption{
Aerodynamic coefficient maps
}
\label{fig:surface_comparison}
\end{figure}

\begin{figure}
\centering
    \begin{adjustbox}{minipage=\linewidth,scale=1.0}
    \begin{subfigure}[t]{0.49\textwidth}
    \centering
    \includegraphics[width=\textwidth]{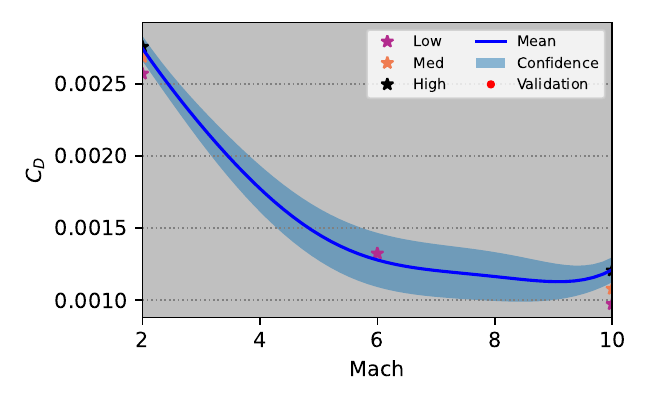}
    \vspace{-8mm}
    \caption{Gradient LMC [AoA=3$^\circ$, Gc=5.0]}
    \end{subfigure}%
    \hfill
    \begin{subfigure}[t]{0.49\textwidth}
    \centering
    \includegraphics[width=\textwidth]{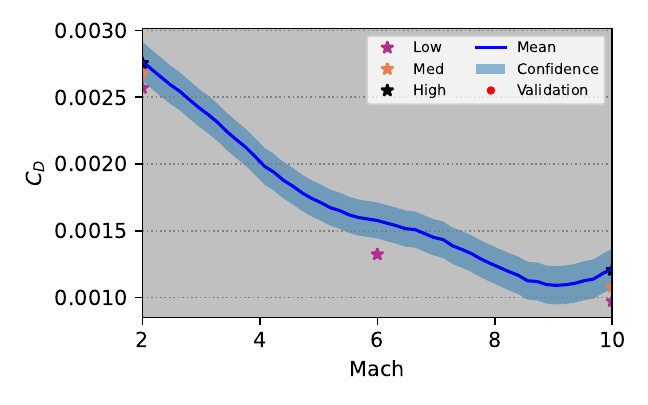}
    \vspace{-8mm}
    \caption{Gradient deep GP (PLL) [AoA=3$^\circ$, Gc=5.0]}
    \end{subfigure}%
    \vfill
    \begin{subfigure}[t]{0.49\textwidth}
    \centering
    \includegraphics[width=\textwidth]{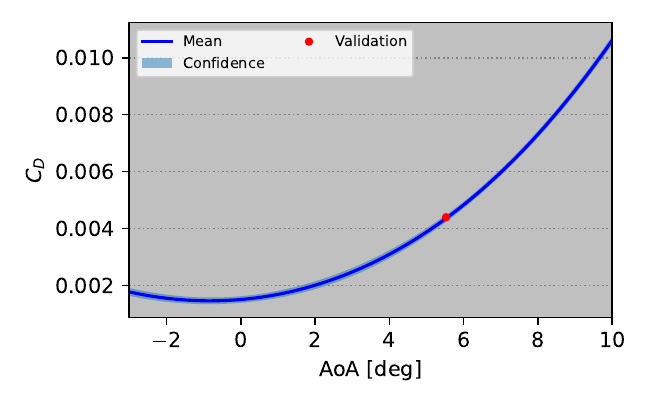}
    \vspace{-8mm}
    \caption{Gradient LMC [Mach=3.495, Gc=5.043]}
    \end{subfigure}
    \hfill
    \begin{subfigure}[t]{0.49\textwidth}
    \centering
    \includegraphics[width=\textwidth]{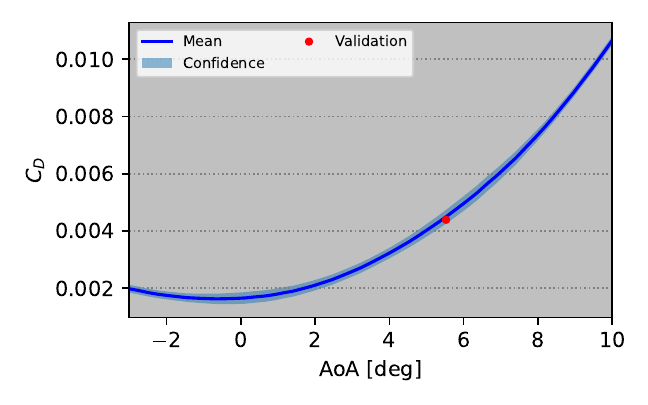}
    \vspace{-8mm}
    \caption{Gradient deep GP (PLL) [Mach=3.495, Gc=5.043]}
    \end{subfigure}
    \vfill
    \begin{subfigure}[t]{0.49\textwidth}
    \centering
    \includegraphics[width=\textwidth]{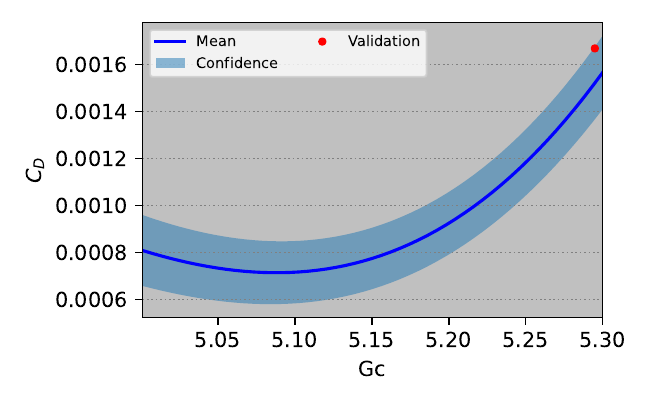}
    \vspace{-8mm}
    \caption{Gradient LMC [Mach=7.744, AoA=-1.51$^\circ$]}
    \end{subfigure}
    \hfill
    \begin{subfigure}[t]{0.49\textwidth}
    \centering
    \includegraphics[width=\textwidth]{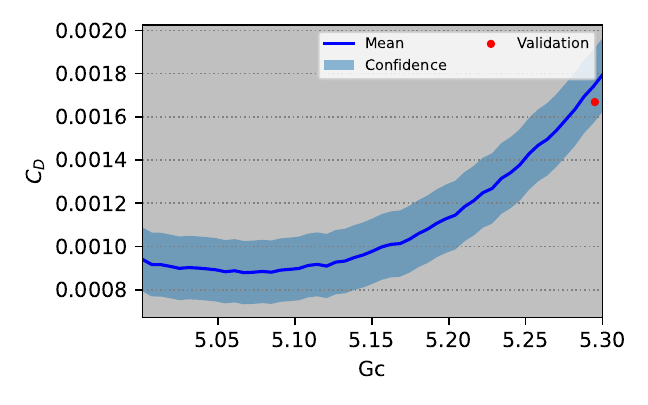}
    \vspace{-8mm}
    \caption{Gradient deep GP (PLL) [Mach=7.744, AoA=-1.51$^\circ$]}
    \end{subfigure}
    \end{adjustbox}{}
\caption{
Comparison of multifidelity GPR methods
}
\label{fig:aerodeck_slice_comparison}
\end{figure}

Fig.\;\ref{fig:surface_comparison} visualizes the aerodynamic coefficient
surfaces as a function of Mach number and angle of attack, with the body curvature
parameter fixed to 5.
This reference surface is generated using a single-fidelity GP model, trained using high
fidelity data at all training and test points.
Fig.\;\ref{fig:aerodeck_slice_comparison} displays three one-dimensional slices
with error bounds for the drag coefficient surface.
Where applicable, training and test data from the high-fidelity
result are also shown.

\subsection{Discussion}
\label{sec:discussion}

The results presented herein show that
(i) deep GPs can be extended to incorporate gradient information and
(ii) that this information is useful for improving their performance in the
multifidelity setting.
For all test cases that we consider, gradient-enhanced deep GPs outperform
gradient-enhanced LMC models.
In particular, the improved performance of gradient-enhanced deep GPs on the aerospace PDE problem
highlights their utility for challenging realistic examples.
The comparative performance advantage of gradient-enhanced deep GP models may further
increase with for problems with higher dimensional inputs, where the restrictive assumptions
that underpin the linear multifidelity models become harder to satisfy, and accurately capturing the input-dependent relationships between different fidelities becomes critical.
Furthermore, training on gradient information enables the GPR models to more accurately 
\emph{predict} gradient information, enabling their deployment when
solving parametric optimization problems, such as shape optimization, codesign, and Bayesian optimization.

However, a major drawback of gradient-enhanced GPs is their increased computational cost when
compared to standard GPs, whose bottleneck results from inverting the covariance matrix of
dimension $(1+d)\times n$. 
This increase in computational cost carries over to the gradient-enhanced deep
GP, requiring inversion of a $(1+d)\times m_\ell$ dimensional covariance
matrix at each layer. 
We have focused on relatively small datasets, which permit the use of
comparatively many inducing points and data points without excessive
computational cost.
Future work should consider how gradient-enhanced deep GPs perform on larger
datasets, which necessitate the use of many less inducing points than data points
and minibatch sampling~\citep{salimbeni2017doubly}.
Scaling issues may be further alleviated by compressing
the gradient-enhanced covariance matrix via directional
derivatives~\citep{padidar2021scaling} or enhanced 
optimization techniques~\citep{hebbal2021bayesian}, which may reduce the number of required training iterations. 

Finally, as discussed in \citep{jankowiak2020parametric, jankowiak2020deep}, the
choice of training objective significantly impacts the predictive distributions
generated by variational GP models.
Most notably, the PLL objective yields significantly richer uncertainty bounds
than the standard ELBO objective, which may be useful in Bayesian or robust
optimization settings.
Future work should extend the gradient-enhancement techniques presented here to
the deep sigma point process~\citep{jankowiak2020deep}, which does not result in
a biased estimator when coupled with the PLL objective.

\section{Conclusion}
\label{sec:conclusion}

This work presented a new deep GP model which incorporates gradient data, then
applied this model in the multifidelity setting.
This capability is particularly relevant when constructing surrogate models for
discretized PDE solvers, where gradient information can be obtained cheaply via
adjoint solutions.
For the examples presented herein, the gradient-enhanced deep GP significantly
outperforms a standard deep GP.
In contrast to existing multifidelity gradient-enhanced methods,
the gradient-enhanced deep GP can handle nonlinear input-dependent relationships
between fidelity levels.
This method leverages sparse VI and can thus be scaled to relatively large
datasets.






\section*{References}

\bibliography{../biblio}
\bibliographystyle{plainnat}

\end{document}